\documentclass{article}

\usepackage{graphicx}%
\usepackage{multirow}%
\usepackage{amsmath,amssymb,amsfonts}%
\usepackage{amsthm}%
\usepackage[title]{appendix}%
\usepackage{xcolor}%
\usepackage{textcomp}%
\usepackage{manyfoot}%
\usepackage{tabularx}
\usepackage{booktabs}%
\usepackage{algorithm}%
\usepackage{algorithmicx}%
\usepackage{algpseudocode}%
\usepackage{listings}%
\usepackage{arxiv}
\usepackage{natbib}

\title{The importance of tensor handling and implementation of physics-informed DeepONets}

\author{
 Nahum Banks \\
  Faculty of Engineering and Physical Sciences\\
  University of Southampton\\
  Southampton, UK\\
  \texttt{nab1g24@soton.ac.uk} \\
   \And
 Gustavo A.M. de Almeida\\
  Faculty of Engineering and Physical Sciences\\
  University of Southampton\\
  Southampton, UK\\
  \And
 Sergio Maldonado\\
  Faculty of Engineering and Physical Sciences\\
  University of Southampton\\
  Southampton, UK\\
  \texttt{s.maldonado@soton.ac.uk} \\
  }

\begin{document}
\maketitle

\begin{abstract}
Physics-informed neural networks (PINNs) reduce or eliminate the need for large labelled datasets by enforcing governing equations during training. Physics-informed Deep Operator Networks (PI-DeepONets) extend this approach to learning solution operators, enabling rapid prediction across families of problems. Despite these advantages, we show that their application to higher-dimensional problems remains limited by the computational cost of training. To address this issue, we propose a modified PI-DeepONet architecture that reduces this cost by removing redundancy in how the Branch network processes input functions across collocation points. The proposed formulation is mathematically equivalent to the original architecture but reduces input tensor sizes and the number of Branch network evaluations. We analyse the computational complexity of both architectures and derive theoretical scaling laws for training time with respect to the number of collocation points and input functions. The proposed architecture is evaluated on 1D, 2D, and 3D linear advection problems. In all cases, it achieves comparable predictive accuracy to the original PI-DeepONet while consistently reducing training time. In 2D experiments, a maximum speed-up of $3.75\times$ is observed, with empirical training-time scaling agreeing with the predicted linear dependence on collocation points and input functions. Furthermore, training a 3D model on an NVIDIA A100 GPU is only feasible using the proposed architecture, as the original formulation exceeds memory limits. These results demonstrate that careful consideration of how PI-DeepONets are implemented can yield substantial gains, enhancing the practical performance of PI-DeepONets and enabling their application to otherwise intractable physics-informed learning problems.
\end{abstract}


\section{Introduction}\label{Introduction}
Scientific Machine Learning (SciML) aims to make use of the vast amounts of prior knowledge that is often available about physical systems when training machine learning models \citep{baker_workshop_2019, rackauckas_universal_2021}. The seminal work by \cite{raissi_physics-informed_2019} introduced Physics-Informed Neural Networks (PINNs), which leverage automatic differentiation to compute derivatives appearing in the governing equations. PINNs encode the underlying Partial Differential Equations (PDEs) governing a physical system, alongside the boundary and initial conditions directly into the loss function, reducing or even eliminating the requirements for labelled training data. Embedding physical constraints in the loss function greatly reduces the acceptable solution space without the need for labelled data \citep{antonion_machine_2024}. In recent years, there has been an abundance of work applying PINNs, ranging from problems in fluid mechanics \citep{mao_physics-informed_2020, cai_physics-informed_2021, de_wolff_assessing_2021, chen_simultaneous_2023, qi_physics-informed_2024, donnelly_physics-informed_2024, brecht_physics-informed_2025}, molecular imaging \citep{awojoyogbe_digital_2024}, solid mechanics \citep{goswami_transfer_2020, haghighat_physics-informed_2022, xu_transfer_2023}, and learning governing equations from data \citep{chen_physics-informed_2021,  yang_b-pinns_2021, xu_transfer_2023, gurevich_learning_2024}, to cite only a few.

Despite their success, a critical limitation for the real-world application of PINNs for forward modelling is that they typically require retraining even for small changes to the domain, initial conditions, or — perhaps most importantly — boundary conditions. This stems from the fact that a conventional PINN takes only the independent variables of the PDE as input into the neural network, while the initial and boundary conditions are incorporated indirectly as soft constraints in the loss function. Consequently, the network is trained to output a solution corresponding to a fixed set of conditions, and does not naturally generalise to different initial or boundary conditions \citep{lu_deepxde_2021, brecht_physics-informed_2025}. As a result, the fast inference times of trained PINNs cannot be easily leveraged across different problem setups, limiting their practical utility in dynamic or rapidly changing scenarios \citep{donnelly_physics-informed_2024}. 

A possible solution for this limitation is to use physics informed neural operators. Neural operators aim to learn mappings between infinite-dimensional function spaces \citep{kovachki_neural_2023}; for example, a mapping from boundary conditions to the PDE's solution across the whole spatiotemporal domain. One notable neural operator framework is the Deep Operator Network (DeepONet) \citep{lu_deeponet_2021}, and its physics-informed variant, the PI-DeepONet \citep{wang_learning_2021}.

DeepONet is grounded in the universal operator approximation theorem, which states that a neural network with a single hidden layer can approximate any non-linear continuous operator $G$ to arbitrary accuracy \citep{tianping_chen_universal_1995}. The learned operator approximation $\hat{G}$ maps an input function $u$ to an output function $\hat{G}(u)$, which can then be evaluated at collocation/query points $y$ to produce $\hat{G}(u)(y)$. \cite{lu_deeponet_2021} extended this framework to use deep neural networks in their unstacked DeepONet. In a DeepONet a single Branch network encodes the input function at $m$ sensor points $\xi_{j\in[1,m]}$, while a single Trunk network encodes the collocation/query point $y=(\mathbf{x},t)$. For example, the input layer of the Branch network could depict a time-dependent boundary condition, whereas the Trunk network functions similarly to a conventional PINN, taking points within the spatiotemporal domain as input. The outputs of these subnetworks are then combined via a dot product to approximate the output function evaluated at the query point. This can be represented by the equation:
\begin{equation}
\hat{G}(u)(y) \approx \sum_{k=1}^s b_k(u(\xi_1),u(\xi_2),...,u(\xi_m))\tau_k(y) ,
\end{equation}
where, $b_k$ are the Branch network outputs, $\tau_k$ are the Trunk network outputs, both of which are of length $s$. The subnetworks require no specific architecture, allowing flexible integration of different networks like feedforward, convolutional, etc. \\

\begin{figure}[!htb]
    \centering
    \includegraphics[height=2.5in]{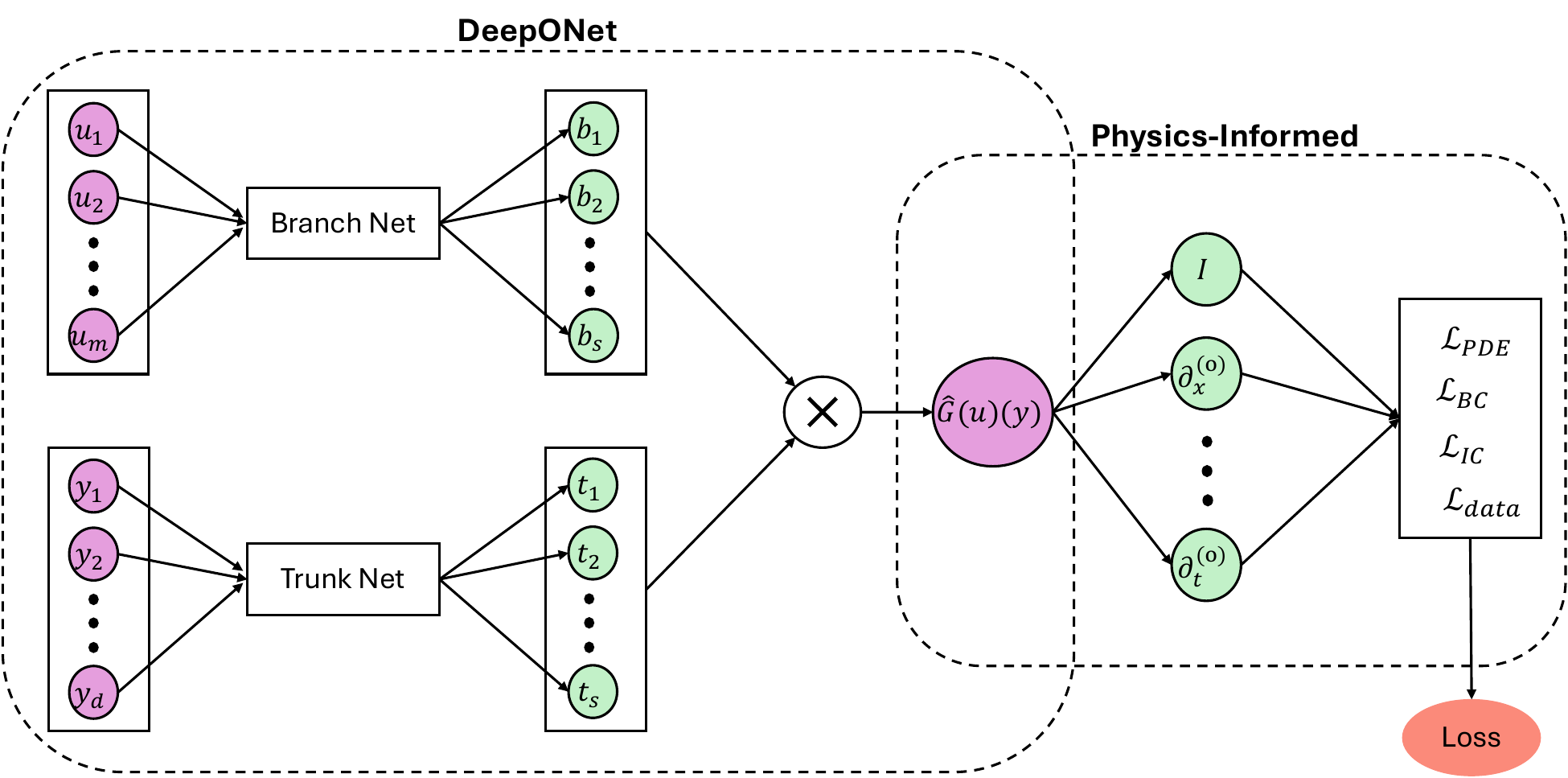}
    \caption{Schematic diagram of a physics-informed Deep Operator Network (PI-DeepONet). $u$ is the input function discretised at $m$ sensor points, and $y$ is the query location for the output function of dimensionality $d$. Derivatives of the output with respect to the input coordinates are calculated via automatic differentiation; these operations are denoted by $\partial_x^{(o)}$ and $\partial_t^{(o)}$, where $o$ is the order of derivative. $\mathcal{L}_{{\text{PDE}}}$, $\mathcal{L}_{\text{BC}}$, $\mathcal{L}_{\text{IC}}$, and $\mathcal{L}_{\text{data}}$ denote the PDE, boundary condition, initial condition, and labelled data loss terms, respectively.}
    \label{Fig:PIDeepONet_Diagram}
\end{figure}

Owing to the simple, feedforward architecture of the subnetworks in the DeepONet, it is straightforward to integrate physics-informed techniques into DeepONet designs. A PI-DeepONet is trained using a loss function formulated in the same way as in PINNs \citep{wang_learning_2021}. A schematic of a PI-DeepONet can be seen in Figure \ref{Fig:PIDeepONet_Diagram}. Derivative terms in the physics loss are calculated by automatic differentiation along the Trunk network, which encodes the collocation points of the output function space. The PI-DeepONet is especially attractive in scenarios where simulations are expensive, high-fidelity data are limited, and adherence to physical laws is critical for model reliability \citep{azizzadenesheli_neural_2024}. Such scenarios include real-time forecasting, uncertainty quantification, and digital twins. This is because the PI-DeepONet can be trained without labelled data, and can then efficiently generate predictions across a variety of problem configurations.

Overall, PI-DeepONet is one of the first physics-informed neural operator frameworks and has received substantial attention since its introduction \citep{wang_learning_2021}, introducing a general approach that allows purely physics-informed operator learning, but broader adoption has been limited compared to data-driven neural operators. This is partly due to the extra computational burden of using automatic differentiation to calculate derivatives in the physics loss terms, which becomes more apparent for high-dimensional systems and when higher order derivatives are present. The high computational cost stems from the fact the number of Trunk evaluations scales exponentially with the number of dimensions (i.e. $N^d$), where $N$ is the is the resolution of each coordinate axis and $d$ is the number of spatiotemporal dimensions \citep{mandl_separable_2025}. As a result, most research on PI-DeepONet to date has focused on low-dimensional problems, using hybrid approaches where physics constraints are combined with labelled data \citep{goswami_physics-informed_2022, goswami_physics-informed_2022-1, eshaghi_variational_2025}.

Recent works have tried to overcome this hurdle by incorporating a separation-of-variables approach to help PINNs and PI-DeepONets scale to higher-dimensional problems \citep{cho_separable_2023, mandl_separable_2025}. In the Separable PI-DeepONet, separate Trunk networks are employed to encode each coordinate axis (i.e. independent variable of the PDE). This PI-DeepONet structure reduces the number of Trunk network evaluations from $N^d$ to $Nd$, making it favourable for high-dimensional problems. However, this method achieves the best performance when the underlying physical system is compatible with a separation-of-variables approach, somewhat limiting the general applicability of the PI-DeepONet framework — one of its most compelling advantages. To the best of the authors’ knowledge, the work by \cite{mandl_separable_2025} is the only so far that has aimed to reduce the computational expense of training PI-DeepONets.

This work investigates an alternative Accelerated Tensor Handling (ATenH) PI-DeepONet architecture which aims to increase its computational efficiency. This architecture takes inspiration from those suggested in \cite{lu_comprehensive_2022}, and applies it in the physics-informed setting. 

The main objectives of this paper are summarised as follows:

\begin{enumerate}
    \item To introduce an alternative PI-DeepONet architecture, hereafter referred to as ATenH, that reduces the computational cost of training and inference.
    \item To derive the theoretical computational speed-ups achieved by ATenH relative to the original PI-DeepONet architecture (Vanilla).
    \item To evaluate the performance of ATenH through a set of increasingly challenging numerical experiments.
    \item To demonstrate that extending PI-DeepONets to higher-dimensional problems presents significant computational challenges, and that architectural improvements such as ATenH are essential for making some problems computationally tractable.
\end{enumerate}
In doing so, we demonstrate that these architectural modifications enable PI-DeepONets to be applied to higher-dimensional PDE systems at greater spatial and temporal resolutions than was previously possible. All numerical results for the Vanilla and ATenH architectures are obtained using PyTorch, demonstrating that these efficiency gains can be realised within standard, widely used, deep learning frameworks.

The remainder of this paper is structured as follows. Section \ref{PI-DeepONet} outlines the PI-DeepONet framework for operator learning and identifies important issues responsible for their limited computational performance; Sections \ref{Case I Architecture} and \ref{Case II Architecture} detail the different PI-DeepONet architectures investigated in this research; Section \ref{sec:scaling_whole_step} derives some theoretical scaling laws for the different architectures; Section \ref{Results} reports training time and accuracy of the Vanilla and ATenH PI-DeepONets for linear advection problems for progressively increasing spatial dimensions; Section \ref{Discussion} provides a discussion of key findings, and Section \ref{Conclusion} provides concluding remarks.

\section*{Notation}
\label{sec:notation}

\begin{table}[ht]
\centering
\caption{Glossary of key variables used throughout this work.}
\label{tab:notation}
\begin{tabular}{ll}
\toprule
Symbol & Description \\
\midrule
$n$ & Number of input functions used during training \\
$p$ & Number of collocation points per input function \\
$m$ & Number of sensor (discretisation) points for input functions \\
$d$ & Dimension of the spatiotemporal domain \\
$s$ & Latent dimension (output size of Branch/Trunk networks) \\
$\mathbf{u}^{(j)}$ & $j$th input function \\
$\xi$ & Sensor point at which input functions are sampled \\
$\mathbf{y} = (\mathbf{x},t)$ & Collocation point in the spatiotemporal domain \\
$\mathbf{x}=(x_1,\dots,x_{d-1})$ & Spatial coordinate \\
$\mathbf{c} = (c_1,\dots,c_{d-1})$ & Advection velocity vector \\
$r_l$ & Number of neurons in layer $l$ (Branch network) \\
$e_l$ & Number of neurons in layer $l$ (Trunk network) \\
$L_B$ & Number of layers in multilayer perceptron (Branch network) \\
$L_T$ & Number of layers in multilayer perceptron (Trunk network) \\
$M$ & Number of Branch networks in PI-DeepONet \\
\bottomrule
\end{tabular}
\end{table}

\section{Methodology}
\label{Methodology}
\subsection{General PI-DeepONet framework}
\label{PI-DeepONet}
The DeepONet is rendered physics-informed (PI-DeepONet) by encoding the PDEs, boundary and initial conditions directly into the loss term, which can take the following form:
\begin{align}
\mathcal{N}[G(u)(\mathbf{x}, t)] &= 0, \quad \mathbf{x} \in \Omega, \quad t \in (0, T] \label{Eq:Governing_Equation} \\
\mathcal{B}[G(u)(\mathbf{x}, t)] &= 0, \quad \mathbf{x} \in \partial \Omega, \quad t \in (0, T] \label{Eq:Boundary_Condition} \\
G(u)(\mathbf{x}, 0) &= v_0(\mathbf{x}), \quad \mathbf{x} \in \Omega \label{Eq:Initial_Condition}
\end{align}
Here, $\mathcal{N}[\cdot]$ represents a general differential operator, $\mathcal{B}[\cdot]$ is a boundary operator, $\mathbf{x}$ is the spatial coordinate, $t$ is the temporal coordinate, $v_0(\mathbf{x})$ defines the initial condition, $\Omega \subset \mathbb{R}^{d-1}$ denotes the spatial domain, and $T$ denotes the upper limit of the temporal domain. 

A composite loss function incorporates each of the conditions:
\begin{align}
\label{loss_equation}
\mathcal{L} =  \mathcal{L}_{\text{PDE}} + \mathcal{L}_{\text{BC}} +  \mathcal{L}_{\text{IC}} +  \mathcal{L}_{\text{data}}.
\end{align}
Here, $\mathcal{L}_{\text{PDE}}$ enforces the differential equation by minimising the residual of $\mathcal{N}[G(u)]$ at collocation points, $(\mathbf{x}_i,t_i)$ inside the spatiotemporal domain, $\mathcal{L}_{\text{BC}}$ penalises violations of boundary conditions at boundary collocation points, and $\mathcal{L}_{\text{IC}}$ enforces the initial condition at initial-time collocation points; $\mathcal{L}_{\text{data}}$ is an optional term that penalises the difference between the neural network's prediction and available observation data. In all of our tests, $\mathcal{L}_{\text{data}}$ is not included, ensuring that training is labelled-data-free.

It is worth examining how input functions and collocation points are provided to the PI-DeepONet. The following architectural descriptions assume that the $j$th function $\mathbf{u}^{(j)}$ (where $1\leq j\leq n$) is discretised at $m$ sensor points, and the corresponding output function is evaluated at $p^{(j)}$ different $d-$dimensional collocation points $\mathbf{y}^{(j)}_{k \in [1,p]}$.

\subsection{Vanilla architecture}
\label{Case I Architecture}
We first introduce the original PI-DeepONet architecture from \cite{wang_learning_2021}. In this architecture, the data triplet (Branch and Trunk network inputs, DeepONet output) for a specific input-output function mapping is given by the following expression:

\[[\mathbf{u},\mathbf{y},G(\mathbf{u})(\mathbf{y})] = 
\left[
\begin{bmatrix}
\mathbf{u}^{(1)} \\
\vdots  \\
\mathbf{u}^{(1)} \\
\mathbf{u}^{(2)} \\
\vdots \\
\mathbf{u}^{(n)} 
\end{bmatrix},
\begin{bmatrix}
\mathbf{y}^{(1)}_1 \\ 
\vdots \\ 
\mathbf{y}^{(1)}_{p^{(1)}} \\ 
\mathbf{y}^{(2)}_1 \\ 
\vdots \\ 
\mathbf{y}^{(n)}_{p^{(n)}}
\end{bmatrix},
\begin{bmatrix}
G(\mathbf{u}^{(1)})(\mathbf{y}^{(1)}_1) \\ 
\vdots \\ 
G(\mathbf{u}^{(1)})(\mathbf{y}^{(1)}_{p^{(1)}}) \\ 
G(\mathbf{u}^{(2)})(\mathbf{y}^{(2)}_1) \\ 
\vdots \\ 
G(\mathbf{u}^{(n)})(\mathbf{y}^{(n)}_{p^{(n)}})
\end{bmatrix}
\right] .
\]

It is important to note that each input function $\mathbf{u}^{(j)}$ is repeated $p^{(j)}$ times to match the number of collocation points it is evaluated at, enabling a pointwise inner product between the Branch and Trunk network outputs. In the most general setting, $p^{(j)}$ may vary across input functions, allowing each input–output mapping to be evaluated on a distinct amount of collocation points. However, in the labelled-data-free setting considered here, it is natural to use the same number of collocation points across all input functions, such that $p^{(j)}=p$ for all $j$. As a result, the Branch network input shape is $(n\times p, m)$, the Trunk network input shape is $(n \times p, d)$, and the PI-DeepONet's final output has shape $(n \times p, 1)$; that is, every output function (which is a mapping of its corresponding input function) evaluated at each of its corresponding collocation points. The Vanilla architecture provides the greatest flexibility, as any number of collocation points can be used with each input function. Pseudocode for the Vanilla architecture can be seen in Algorithm~\ref{alg:Arch1}.

\begin{algorithm}
\caption{Vanilla PI-DeepONet Training}
\label{alg:Arch1}
\begin{algorithmic}[1]
\Require Input functions $\{\mathbf{u}^{(j)}\}_{j=1}^n$, collocation points $\{\mathbf{y}_k\}_{k=1}^p$
\Ensure Network outputs $\mathbf{G}(\mathbf{u})(\mathbf{y}) \in \mathbb{R}^{n p}$

\State Repeat each input function $\mathbf{u}^{(j)}$ for all collocation points
\Statex $\text{repeat}(\mathbf{u}^{(j)}, p)\rightarrow \mathbf{U} \in \mathbb{R}^{(n p) \times m}$

\State Tile collocation points for all input functions
\Statex $\text{tile}(\mathbf{y}_k, n) \rightarrow \mathbf{Y} \in \mathbb{R}^{(n p) \times d}$

\State Compute Branch network output
\Statex $\mathcal{B}(\mathbf{U}) \rightarrow \mathbf{B} \in  \mathbb{R}^{(n p) \times s}$

\State Compute Trunk network output
\Statex $\mathcal{T}(\mathbf{Y}) \rightarrow \mathbf{T} \in  \mathbb{R}^{(n p) \times s}$

\State Compute final output via inner product
\Statex $G_{i} = \sum_{\ell=1}^{s} B_{i\ell} \, T_{i\ell}$

\State Evaluate physics-informed loss 
\Statex
$\mathcal{L} =  \mathcal{L}_{\text{PDE}} + \mathcal{L}_{\text{BC}} +  \mathcal{L}_{\text{IC}}$

\State Update model parameters via gradient-based optimisation

\end{algorithmic}
\end{algorithm}

Next, we introduce an alternative architecture, ATenH (Accelerated Tensor Handling), which contains a modification that reduces the number of operations and the dataset sizes involved in a PI-DeepONet evaluation. 

\subsection{Accelerated Tensor Handling (ATenH) architecture}
\label{Case II Architecture}
In the proposed alternative architecture, the number of Branch passes is reduced by a factor $p$ by re-using the output of a single Branch network pass for all corresponding collocation points. In particular, the input size of the Branch network becomes $(n, m)$ and the output of the Branch network becomes $(n, s)$, which contrasts with $(n\times p, m) \to (n\times p, s)$ for the Vanilla architecture. Additionally, the input data for the Trunk network is reshaped as $(n, p, d)$ – rather than $(n \times p, d)$ as in the Vanilla architecture. This formulation requires a constant amount of collocation points across all input functions, a constraint not imposed by the Vanilla version. The data triplet for a specific input-output function mapping is given by:

\begin{align*}
[\mathbf{u},\mathbf{y},G(\mathbf{u})(\mathbf{y})]
&=
\left[
\begin{bmatrix}
\mathbf{u}^{(1)}\\
\mathbf{u}^{(2)}\\
\vdots\\
\mathbf{u}^{(n)}
\end{bmatrix},
\begin{bmatrix}
\mathbf{y}^{(1)}_1 & \mathbf{y}^{(1)}_2 & \cdots & \mathbf{y}^{(1)}_p\\
\mathbf{y}^{(2)}_1 & \mathbf{y}^{(2)}_2 & \cdots & \mathbf{y}^{(2)}_p\\
\vdots & \vdots & & \vdots\\
\mathbf{y}^{(n)}_1 & \mathbf{y}^{(n)}_2 & \cdots & \mathbf{y}^{(n)}_p
\end{bmatrix},
\right.
\\[1ex]
&\qquad
\left.
\begin{bmatrix}
G(\mathbf{u}^{(1)})(\mathbf{y}^{(1)}_1) & G(\mathbf{u}^{(1)})(\mathbf{y}^{(1)}_2) & \dots & G(\mathbf{u}^{(1)})(\mathbf{y}^{(1)}_p) \\ 
G(\mathbf{u}^{(2)})(\mathbf{y}^{(2)}_1) & G(\mathbf{u}^{(2)})(\mathbf{y}^{(2)}_2) & \dots & G(\mathbf{u}^{(2)}(\mathbf{y}^{(2)}_p)\\ 
\vdots & \vdots & & \vdots\\ 
G(\mathbf{u}^{(n)})(\mathbf{y}^{(n)}_1) & G(\mathbf{u}^{(n)})(\mathbf{y}^{(n)}_2) & \dots & G(\mathbf{u}^{(n)})(\mathbf{y}^{(n)}_p) \\
\end{bmatrix}
\right].
\end{align*}

It is important to note that the input function $\mathbf{u}^{(j)}$ is no longer repeated $p$ times, as was the case with the Vanilla architecture. As a consequence, the inner product cannot be performed in the same manner due to the subnetworks' output shapes no longer matching. Thus, the Branch network's output must be broadcasted beforehand. This operation can be performed efficiently in many Python libraries using an einsum (Einstein summation) operation. This results in the PI-DeepONet output having the desired shape $(n, p, 1)$, which may be trivially converted to the shape $(n \times p, 1)$ of the Vanilla architecture. Pseudocode for ATenH can be seen in Algorithm~\ref{alg:Arch2}. A summary of the data shapes for the different architectures can be seen in Table~\ref{Table:CaseIvsCaseII_DataShapes}.

\begin{algorithm}
\caption{ATenH (new proposal) PI-DeepONet Training}
\label{alg:Arch2}
\begin{algorithmic}[1]
\Require Input functions $\{\mathbf{u}^{(j)}\}_{j=1}^n$, collocation points $\{\mathbf{y}_k\}_{k=1}^p$
\Ensure Network outputs $\mathbf{G}(\mathbf{u})(\mathbf{y}) \in \mathbb{R}^{n \times p}$

\State No need to repeat each input function $\mathbf{u}^{(j)}$ for all collocation points, simply
\Statex $\mathbf{u}^{(j)}\rightarrow \mathbf{U} \in \mathbb{R}^{n \times m}$

\State Tile collocation points for all input functions
\Statex $\text{tile}(\mathbf{y}_k, n) \rightarrow \mathbf{Y} \in \mathbb{R}^{n \times p \times d}$

\State Compute Branch network output
\Statex $\mathcal{B}(\mathbf{U}) \rightarrow \mathbf{B} \in  \mathbb{R}^{n \times s}$

\State Compute Trunk network output
\Statex $\mathcal{T}(\mathbf{Y}) \rightarrow \mathbf{T} \in  \mathbb{R}^{n \times p \times s}$

\State Compute final output via dot product over latent dimension
\Statex $G_{j k} = \sum_{\ell=1}^{s} B_{j \ell} \, T_{j k \ell}$

\State Evaluate physics-informed loss 
\Statex
$\mathcal{L} =  \mathcal{L}_{\text{PDE}} + \mathcal{L}_{\text{BC}} +  \mathcal{L}_{\text{IC}}$

\State Update model parameters via gradient-based optimisation

\end{algorithmic}
\end{algorithm}

\begin{table}[ht]
\centering
\begin{tabularx}{\textwidth}{l c c c c X}
\toprule
Architecture & Branch input & Trunk input & Branch output & Trunk output & Combination method \\
\midrule
Vanilla   & $(n p, m)$ & $(n p, d)$ & $(n p, s)$ & $(n p, s)$ & Inner product \\
ATenH  & $(n, m)$   & $(n, p, d)$ & $(n, s)$   & $(n, p, s)$ & Inner product (with broadcasting) \\
\bottomrule
\end{tabularx}
\caption{A comparison of data processing choices in the two PI-DeepONet architectures herein presented. Here, $m$ is the number of discretisation points for the input functions (the number of sensor points), $d$ is the physical dimensionality of the collocation points, $n$ is the number of input functions, $p$ is the number of collocation points per function, and $s$ is the length of the output layer of the (Branch and Trunk) subnetworks}
\label{Table:CaseIvsCaseII_DataShapes}
\end{table}

\subsection{Theoretical scaling laws for a full training step}
\label{sec:scaling_whole_step}
We now analyse the computational cost of the different architectures over a training step, including automatic differentiation for first-order physics derivatives and gradients with respect to model parameters. The cost of computing the loss at the boundaries, where no physics derivatives are calculated (we only use Dirichlet boundary conditions), is not included, since the number of boundary collocation points is generally much less than the number of interior points used for calculating the physics loss.

\subsubsection{Cost model}
Throughout this section, computational cost is measured in units proportional to the number of input-weight products evaluated; i.e. scalar multiplications of the form $w_i x_i$, where $w_i$ and $x_i$ are weights and inputs, respectively. This choice highlights the dominant scaling behaviour.

For a neuron in layer $l$ with $r_{l-1}$ inputs, the activation is given by:
\begin{equation}
\sigma\!\left(\sum_{i=1}^{r_{l-1}} w_i x_i + \beta\right),
\end{equation}
where $\sigma(\,)$ is the activation function and $\beta$ is the neuron's bias. Assuming the computational cost is dominated by the weighted sum and treating the cost of the activation function as a constant, evaluating this neuron processes $r_{l-1}$ weighted inputs and therefore has a cost proportional to $r_{l-1}$.
Bias additions and application of activation functions are pointwise scalar operations, which we therefore neglect.

A fully connected layer with $r_{l-1}$ inputs and $r_l$ neurons therefore has a cost proportional to:
\begin{equation}
r_{l-1} r_l.    
\end{equation}

\paragraph{Forward pass cost}
For an $L$-layer MLP, the total cost of a forward pass is
\begin{equation}
C_{\mathrm{MLP}} \sim \sum_{l=1}^{L} r_{l-1} r_l ,
\end{equation}
where $r_0$ denotes the size of the MLP's input layer.

This cost model is equivalent, up to a constant factor, to the conventional floating-point operation (FLOP) count. Each weighted input contributes one multiplication and one addition, which is commonly implemented as a fused multiply-add (FMA) instruction.

\paragraph{Dot product}
The DeepONet output is formed by the dot product
\begin{equation}
\sum_{i=1}^{s} b_i \tau_i ,    
\end{equation}
which processes $s$ weighted inputs, where $b_i$ and $\tau_i$ are the Branch and Trunk network outputs, respectively. Thus, in accordance with this cost model, a single dot product has a cost proportional to $s$. A dot product is performed for every input functions and collocation point pairing, meaning the total cost is
\begin{equation}
C_{\mathrm{dot}} \sim n p s .
\end{equation}

\paragraph{Backward pass cost}
Reverse-mode automatic differentiation for a standard MLP involves two dominant operations per layer:
\begin{enumerate}
    \item propagation of the error signal via matrix-vector multiplication, and
    \item computation of weight gradients via matrix multiplication.
\end{enumerate}
Both operations process $r_{l-1}r_l$ weighted products, and a summation over inputs per layer. As a result, both operations have the same asymptotic cost when measured in units proportional to the number of weighted interactions, and each are roughly equal to the total of input-weight products performed during the forward pass. Therefore, the cost of computing gradients with respect to model parameters is approximately
\begin{equation}
C_{\mathrm{backward}} \approx 2 C_{\mathrm{forward}} .
\end{equation}

This approximation neglects element-wise operations (Hadamard products and activation function derivatives),
which contribute lower cost.

When automatic differentiation is used to compute physics derivatives, the computational graph is extended. The same backward-pass reasoning applies, but not every layer necessarily performs both dominant operations. In particular, when differentiating with respect to input or coordinate variables (rather than model parameters), the computation of weight gradients is not required. These distinctions are accounted for explicitly below.

\subsubsection{Vanilla architecture}
The Vanilla architecture uses Branch network inputs of shape $[n p, m]$ and Trunk network inputs of shape $[n p, d]$.

The cost of a forward pass through the Branch network is
\begin{equation}
C_{\mathrm{branch}} \sim n p \sum_{l=1}^{L_B} r_{l-1} r_l ,
\end{equation}
and for the Trunk network,
\begin{equation}
C_{\mathrm{trunk}} \sim n p \sum_{l=1}^{L_T} e_{l-1} e_l,
\end{equation}
where $r_l$ and $e_l$ denote the number of neurons in the $l$th layer of the Branch and Trunk networks, respectively, and $L_B$ and $L_T$ denote the corresponding number of layers. The Branch network has input dimension $r_0 = m$, and the Trunk network has input dimension $e_0 = d$.

The dot product contributes
\begin{equation}
C_{\mathrm{dot}} \sim n p s .
\end{equation}

\paragraph{Physics derivatives}
First derivatives treat the Branch network outputs as constants. Backpropagation therefore traverses only the Trunk network and the dot product. The cost of computing physics derivatives is
\begin{equation}
C_{u_x} \sim n p \sum_{l=1}^{L_T} (e_{l-1} e_l) + n p s .
\end{equation}

\paragraph{Loss gradients}
The loss depends on both $u$ and $u_x$. Backpropagation therefore traverses the entire computational graph. The Trunk network appears twice (forward and derivative paths), while the Branch network appears once.
Applying the $2\times$ backward rule gives
\begin{equation}
C_{\mathrm{loss}}
\sim 2 \left(
2 n p \sum_{l=1}^{L_T} e_{l-1} e_l
+ M n p \sum_{l=1}^{L_B} r_{l-1} r_l
\right)
+ 2 n p s .
\end{equation}
where $M$ is the number of Branch networks, which is relevant for problems with multiple input functions \citep{jin_mionet_2022}.

\paragraph{Total cost}
Summing all contributions yields
\begin{align}
C_{\mathrm{V}}
&\sim M n p \sum_{l=1}^{L_B} r_{l-1} r_l
+ 2 n p \sum_{l=1}^{L_T} e_{l-1} e_l
+ 2 \times \left(
2 n p \sum_{l=1}^{L_T} e_{l-1} e_l
+ M n p \sum_{l=1}^{L_B} r_{l-1} r_l
\right)
+ 4 n p s, \nonumber \\
C_{\mathrm{V}}
&\sim 3 M n p \sum_{l=1}^{L_B} r_{l-1} r_l
+ 6 n p \sum_{l=1}^{L_T} e_{l-1} e_l
+ 4 n p s. \label{eq:CaseI_TotalCost}
\end{align}

Factoring out the number of collocation points $p$, and the number of input functions, $n$,
\begin{align}
C_{\mathrm{V}}
&\sim np \Big(3M\sum_{l=1}^{L_B} r_{l-1} r_l + 6\sum_{l=1}^{L_T} e_{l-1} e_l + 4s\Big). \label{eq:CaseI_Factoredp_1}
\end{align}

Using the simplifications introduced in Section~\ref{Numerical Methodology}, namely $r_0=m$, $e_0=d$, $L_B=L_T=L$, and $e_{l\geq 1}=r_{l\geq 1}=e$, Equation~\ref{eq:CaseI_Factoredp_1} becomes
\begin{align}
C_{\mathrm{V}}
&\sim np\left((3Mm + 6d)e + (L-1)(3M+6)e^2 + 4s\right). \label{eq:CaseI_Factoredp_2}
\end{align}

\subsubsection{ATenH architecture}
The ATenH architecture uses Branch network inputs of shape $[n, m]$ and Trunk network inputs of shape $[n, p, d]$, such that

\begin{align}
C_{\mathrm{branch}} &\sim n \sum_{l=1}^{L_B} r_{l-1} r_l , \\
C_{\mathrm{trunk}}  &\sim n p \sum_{l=1}^{L_T} e_{l-1} e_l , \\
C_{\mathrm{dot}} &\sim n p s .
\end{align}

Physics derivatives and loss backpropagation follow the same structure as the Vanilla architecture, yielding
\begin{equation}
C_{\mathrm{ATenH}} \sim 3Mn\sum_{l=1}^{L_B} r_{l-1}r_l + 6np\sum_{l=1}^{L_T} e_{l-1}e_l + 4nps. \label{eq:CaseII_TotalCost}
\end{equation}

Factoring out the number of collocation points $p$, and the number of input functions, $n$,
\begin{align}
C_{\mathrm{ATenH}}
&\sim n\left[3M\sum_{l=1}^{L_B} r_{l-1} r_l + 2p\left(3\sum_{l=1}^{L_T} e_{l-1} e_l + 2s\right) \right]. \label{eq:CaseII_Factoredn_1}
\end{align}

Using the simplifications introduced in Section~\ref{Numerical Methodology}, namely $r_0=m$, $e_0=d$, $L_B=L_T=L$, and $e_{l\geq 1}=r_{l\geq 1}=e$, Equation~\ref{eq:CaseII_Factoredn_1} becomes
\begin{align}
C_{\mathrm{ATenH}}
&\sim n\left[3Mme + 3M(L-1)e^2 + 2p\left(3de + 3(L-1)e^2 + 2s\right)\right].
\label{eq:CaseII_Factoredn_2}
\end{align}

\subsubsection{Theoretical Speed-up}
The previous derivations allow us to estimate a theoretical speed-up factor, $\phi \sim C_\text{V}/C_\text{ATenH}$,  between the optimised PI-DeepONet architecture (ATenH) and the original architecture (Vanilla).
\begin{align}
    \phi &\sim \frac{3 M p \sum_{l=1}^{L_B} r_{l-1} r_l + 6 p \sum_{l=1}^{L_T} e_{l-1} e_l + 4 p s}{3M\sum_{l=1}^{L_B} r_{l-1}r_l + 6p\sum_{l=1}^{L_T} e_{l-1}e_l + 4ps} \label{eq:trainingtime_scaling_1}
\end{align}
With the simplifications from Section~\ref{Numerical Methodology}, Equation~\ref{eq:trainingtime_scaling_1} becomes
\begin{align}
    \phi &\sim \frac{3Mmpe + 6pde + 3p(M+2)(L-1)e^2 + 4ps}{3Mme + 6pde + 3(M+2p)(L-1)e^2 + 4ps} \label{eq:trainingtime_scaling_2}
\end{align}

The derived speed-up factor is independent of $n$, and reaches a constant (with all other parameters fixed) for a sufficiently large $p$.

\subsection{Numerical tests}
\label{Numerical Methodology}
The numerical experiments in the following section compare the performance of the different PI-DeepONet architectures, focusing on training time and how this scales with key hyper parameters and spatial dimensionality. Since the two architectures share the same mathematical operations – the difference being exclusively in data handling – the optimal parameters should be common to both. Ensuring a fair and consistent hyperparameter selection is therefore essential, especially when assessing scaling with problem dimensionality.

Tests are based on the solution of the linear advection equation in one, two and three spatial dimensions, which provides progressively more stringent computational and memory demands, which is governed by the following PDE:
\begin{align}
    \frac{\partial f}{\partial t}(\mathbf{x}, t) + \mathbf{c} \cdot \nabla f(\mathbf{x}, t) = 0, 
    \quad \forall (\mathbf{x}, t) \in [0,1]^{d-1} \times (0,1],
\end{align}
where $f$ is the advected field, $(\mathbf{x}, t)$ denotes the spatial and temporal coordinates, and $\mathbf{c} = (c_1, \dots, c_{d-1})$ is the constant advection velocity in each spatial dimension.

The problems are constrained by the following initial and boundary conditions:
\begin{align}
    f(\mathbf{x}, 0) &= 0, \\
    f(\mathbf{x}, t)\big|_{x_i = 0} &= 
    A_i \exp\left(-\frac{(t - 0.5)^2}{2\sigma_t^2}\right)
    \prod_{\substack{j=1 \\ j \ne i}}^{d-1}
    \exp\left(-\frac{(x_j - 0.5)^2}{2\sigma_j^2}\right),
    \quad i = 1, \dots, d-1,
\end{align}
where $A_i$ denotes the amplitude associated with the boundary $x_i = 0$, $\sigma_t$ is the temporal standard deviation of the Gaussian profile, and $\sigma_j$ is the standard deviation in the $j$th spatial dimension.

The exact solutions can be expressed in a unified form using the method of characteristics. Define the backtracking times, $\mathcal{S}_i$,
\begin{align}
    \mathcal{S}_i = t - \frac{x_i}{c_i}, \quad i = 1, \dots, d-1,
\end{align}
which correspond to the time at which the characteristic passing through $(\mathbf{x}, t)$ intersects the boundary $x_i = 0$. The solution is then given by
\begin{align}
f(\mathbf{x}, t) =
\begin{cases}
0, & \text{if } \max_i \mathcal{S}_i < 0, \\
f_{i^*}(\mathbf{x}, t), & \text{otherwise},
\end{cases}
\end{align}
where
\begin{align}
i^* = \arg\max_{1 \le i \le d-1} \mathcal{S}_i,
\end{align}
and the contribution from each boundary $x_i=0$ is defined as
\begin{align}
f_i(\mathbf{x}, t) =
A_i \exp\left(-\frac{(\mathcal{S}_i - 0.5)^2}{2\sigma_t^2}\right)
\prod_{\substack{j=1 \\ j \ne i}}^{d-1}
\exp\left(-\frac{\left(x_j - 0.5 - c_j \frac{x_i}{c_i}\right)^2}{2\sigma_j^2}\right),
\quad i=1,\dots,d-1.
\end{align}
The solution $f_{i^*}$ is obtained by selecting the boundary contribution corresponding to the characteristic with the largest backtracking time.

We fix $c_i=1$ for all $i$, $\sigma_t=\sigma_j=0.07$, and allow $A_i$ to vary independently, producing a range of boundary conditions the PI-DeepONet must generalise to. Specifically, $A_i$ is randomly sampled from the uniform distribution $[0.1,1]$ yielding $n$ distinct boundary condition realisations which are fed to the Branch network during training. Although the model is therefore restricted to Gaussian boundary conditions and cannot extrapolate to qualitatively different boundary-condition families, this is appropriate for the purpose of evaluating computational performance.

\begin{figure}[!htb]
    \centering

    \begin{minipage}[t]{0.49\textwidth}
        \centering
        \includegraphics[width=\linewidth]{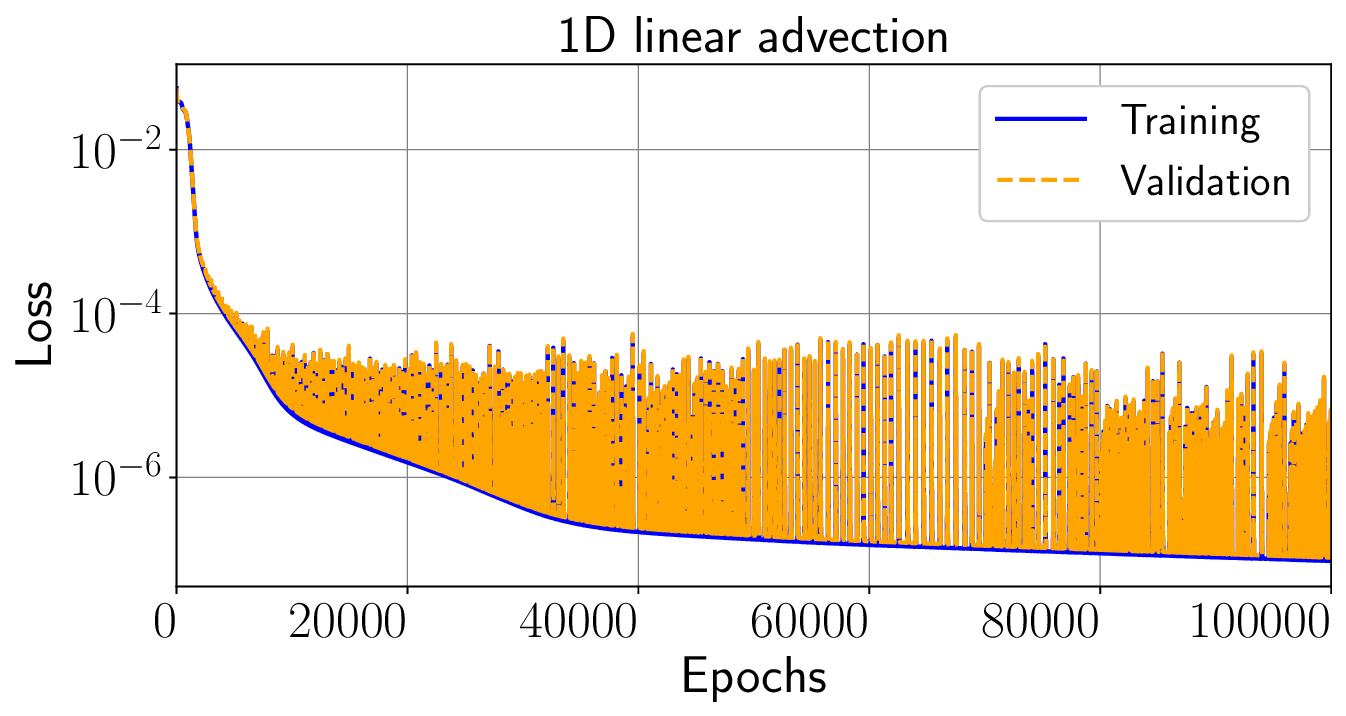}
    \end{minipage}
    \begin{minipage}[t]{0.49\textwidth}
        \centering
        \includegraphics[width=\linewidth]{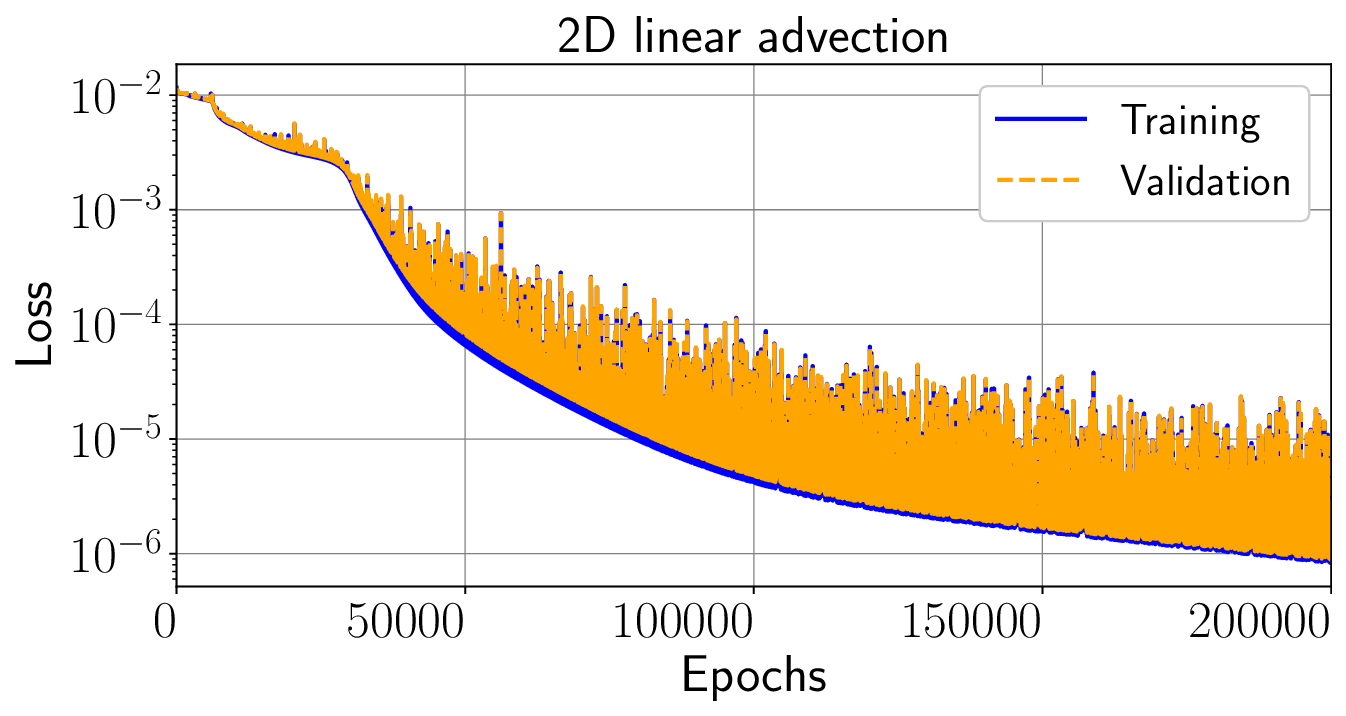}
    \end{minipage}
    \caption{Training and validation loss profiles as a function of training epoch for 1D (left) and 2D (right) linear advection, trained using the ATenH PI-DeepONet. The training and validation loss curves remain closely aligned throughout training indicating no overfitting. The rate of decrease in the loss begins to plateau after approximately 50,000 epochs for 1D linear advection and 100,000 epochs for 2D linear advection}
    \label{fig:Val_Loss}
\end{figure}

The number of training epochs for each problem was determined via preliminary tests which included validation-loss monitoring for the 1D and 2D advection problems: the validation loss was computed at every epoch (without contributing to weight updates) and compared with the training loss to detect overfitting and identify the onset of loss plateauing, indicating diminishing returns from further training. The choice of epochs was based on observing no overfitting, and achieving a low loss, as seen in Fig~\ref{fig:Val_Loss}. The same number of epochs was used for the Vanilla and ATenH architectures to allow for a fair comparison. 50,000 epochs, 100,000 epochs, and 200,000 epochs were selected for 1D, 2D, and 3D linear advection problems, respectively.

For all numerical experiments, collocation points are sampled independently from a uniform distribution over the computational domain. The number of collocation points was chosen so that the average spacing between randomly sampled points remained constant across dimensions, according to $p = \left(\frac{\ell}{\Delta x}\right)^d$, where $\ell=1$ denotes the characteristic length of the domain in each dimension, and $\Delta x$ is the target average spacing between points. Accordingly, we set 10 collocation points per spatiotemporal dimension, since increasing this further (say, by a factor of 2) produced comparable accuracy while causing memory issues, thus preventing the Vanilla architecture from successfully completing beyond the 1D advection case. Failure to complete the relatively simple task of solving the 2D linear advection problem discretized with 20 points along each dimension sheds light onto the need for improved faster methods. In practice, separate collocation points must be sampled for evaluating the loss from the boundary conditions, the initial condition and physics loss. We denote their amounts as $p_{b_{x_i}}$ for $i=1,...,d-1$, $p_0$ and $p_I$, respectively. This dimensional scaling yielded higher predictive accuracy in 2D and 3D than alternative sampling strategies tested.

We use $n=500$ training input functions and $n_{\text{test}}=50$ testing input functions, consistent with prior PI-DeepONet studies (e.g. \citealp{goswami_physics-informed_2022, keppler_learning_2025, mandl_separable_2025}), which is sufficient to ensure generalisation. More complex problems may require larger datasets; for instance, \cite{brecht_physics-informed_2025} employ $\sim 10^4$ input functions. Each input function was discretised using $m=50^{(d-1)}$ sensor points per (spatiotemporal) dimension. Although we consider Gaussian inputs with varying amplitudes, this resolution reflects more general settings where inputs may contain finer-scale variations while keeping training costs manageable. 

In the 2D and 3D linear advection problems there are multiple input functions, corresponding to the inflow boundaries $x_i = 0$ for $i = 1, \dots, d-1$. Each boundary condition is parametrised by an independent amplitude $A_i$, and therefore defines a distinct input function. To handle the multiple independent boundary condition inputs, we employ the Multiple-Input DeepONet (MIONet) structure \cite{jin_mionet_2022}, which uses separate Branch networks to process each input function. In all our experiments, both Branch networks adopt either the Vanilla or the ATenH architecture. The outputs from the separate Branch networks and the Trunk network are then combined via an inner product across all Branch networks and the Trunk network. As an illustrative example, in the 2D linear advection problem with spatial coordinates $(x_1, x_2)$, there are two boundary inputs corresponding to $x_1 = 0$ and $x_2 = 0$. For $n=500$ samples, each boundary condition is discretised using $50$ points per spatial and temporal dimension, yielding $m = 50^2 = 2500$ input degrees of freedom per boundary. The boundary and initial condition losses are evaluated at $10^2$ collocation points each, while the interior physics loss is evaluated at $10^3$ collocation points.

\begin{figure}[!htb]
    \centering

    \begin{minipage}[t]{0.49\textwidth}
        \centering
        \includegraphics[width=\linewidth]{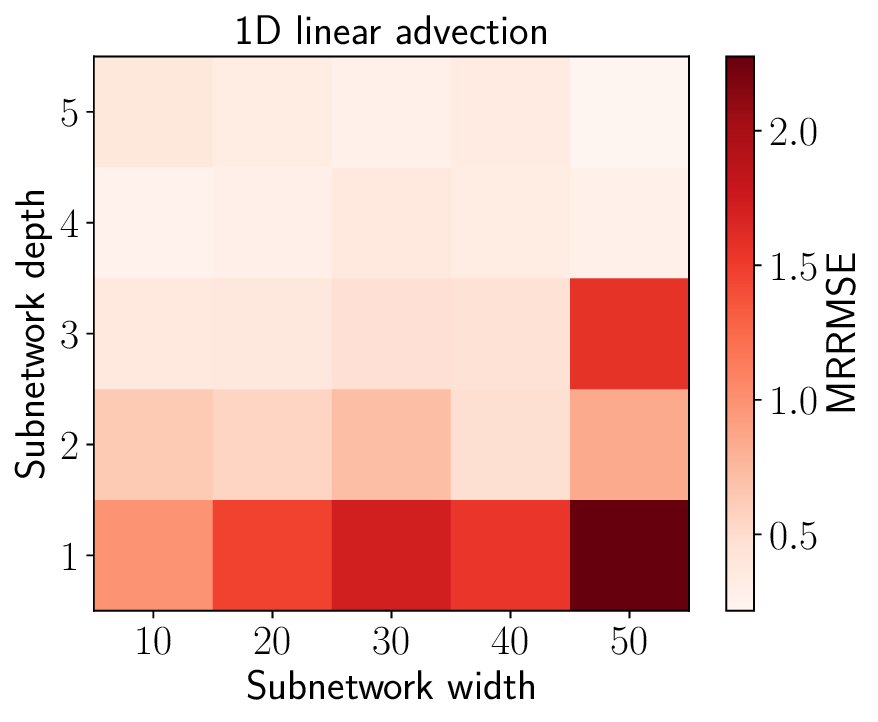}
    \end{minipage}
    \hfill
    \begin{minipage}[t]{0.49\textwidth}
        \centering
        \includegraphics[width=\linewidth]{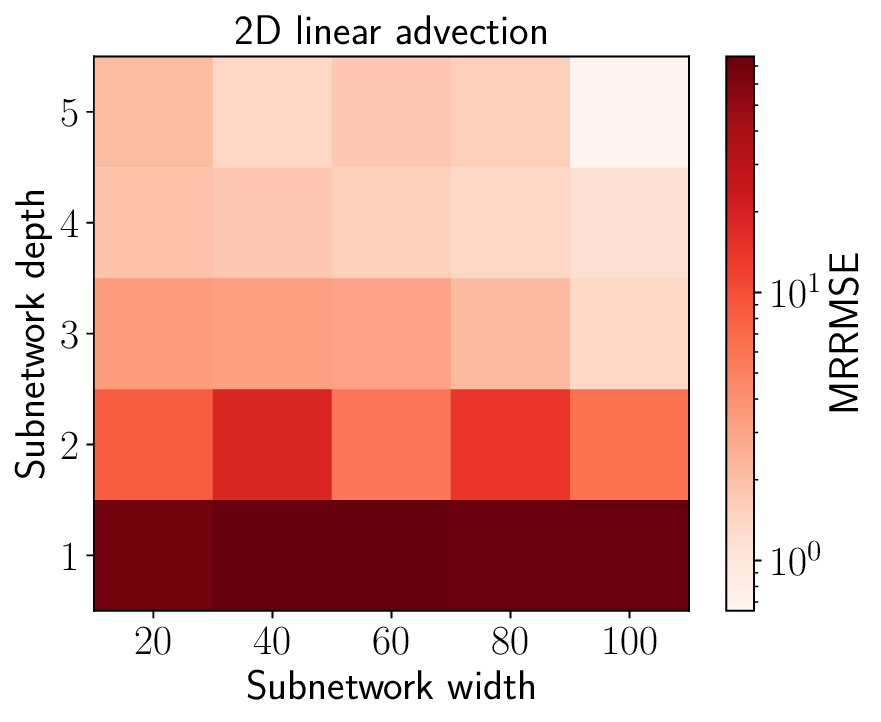}
    \end{minipage}
    \caption{Grid searches exploring the depth and width of the subnetworks for (left) 1D linear advection and (right) 2D linear advection. The same width and depth was used across Branch and Trunk networks to reduce the complexity of the hyperparameter space, alongside keeping the layer widths constant across hidden layers}
    \label{fig:grid_search}
\end{figure}

Subnetwork depths and widths were selected via grid searches on the 1D and 2D advection problems, which can be seen in Fig~\ref{fig:grid_search}. To avoid overcomplicating the hyperparameter space, both subnetworks used the same width and depth. In both cases, four hidden layers provided the best compromise between computational cost and accuracy. Similarly, we select widths of 40 neurons per hidden layer for 1D and 60 for 2D. A dedicated grid search for 3D advection was not performed due to the prohibitive computational cost; instead, we heuristically inferred subnetwork widths from the lower-dimensional results, setting 80 neurons for the 3D problem. All hidden and output layers were kept at the same width to avoid unnecessarily complication of the hyperparameter space. A large proportion of computational resources was not devoted to extensive hyperparameter optimisation, as the primary objective of this study is to assess the computational performance of different architectures rather than to maximise predictive accuracy. The configurations used were sufficient to reliably learn the solutions of all test problems, as demonstrated in the following section.

All models were trained in PyTorch using the Adam optimiser \citep{kingma_adam_2014}, with a learning rate of $10^{-4}$ and $\tanh$ activation functions, on an NVIDIA A100 GPU.

To quantify the predictive accuracy of the architectures, we use the mean relative root mean squared error (MRRMSE), defined as
\begin{equation}
\mathrm{MRRMSE} = \frac{1}{n_{\text{test}}} \sum_{i=1}^{n_{\text{test}}}
\frac{\lVert \hat{u}_i - u_i \rVert_2}{\lVert u_i \rVert_2} \times 100\%,
\end{equation}
where $u_i$ denotes the exact solution associated with the $i$th input function, $\hat{u}_i$ is the corresponding model prediction, and $\lVert \cdot \rVert_2$ denotes the discrete $L^2$ norm over the spatiotemporal evaluation grid. MRRMSE is only computed in regions of the domain where $u_i \neq 0$.

\section{Results}
\label{Results}
\subsection{1D Linear Advection}
\label{1D Linear Advection}
\begin{figure}[!htb]
    \centering

    \begin{tabular}{@{}c@{}c@{}c@{}}
        \includegraphics[width=0.33\textwidth]{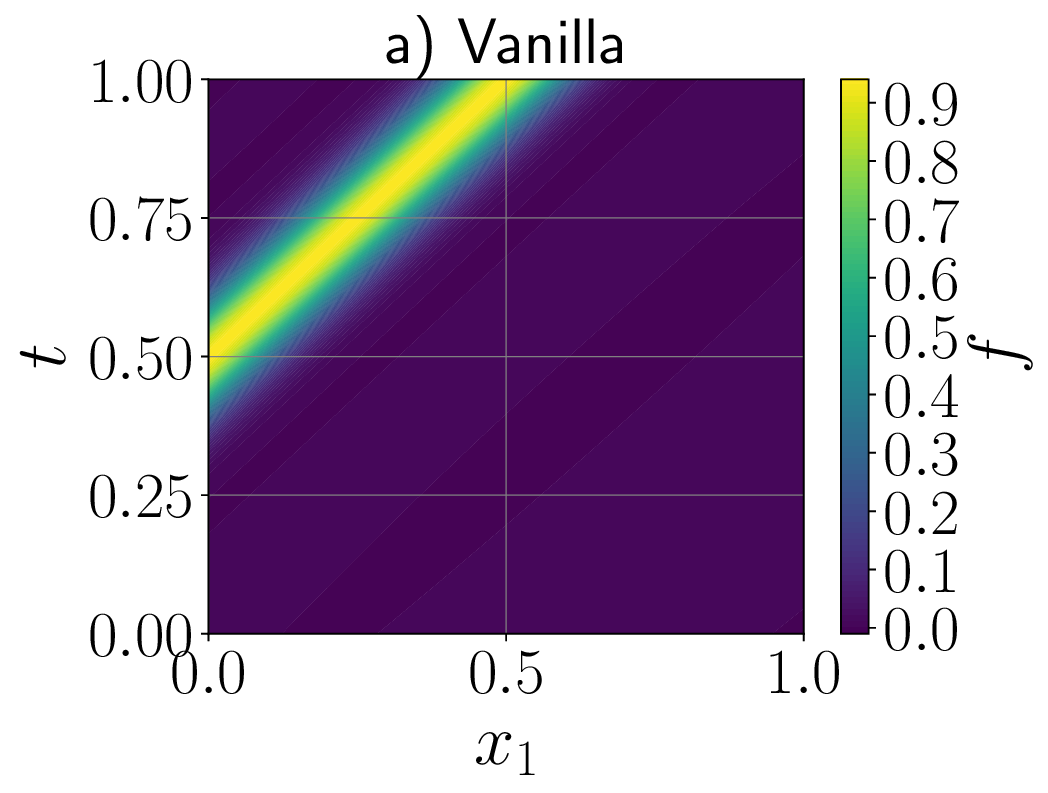} &
        \includegraphics[width=0.33\textwidth]{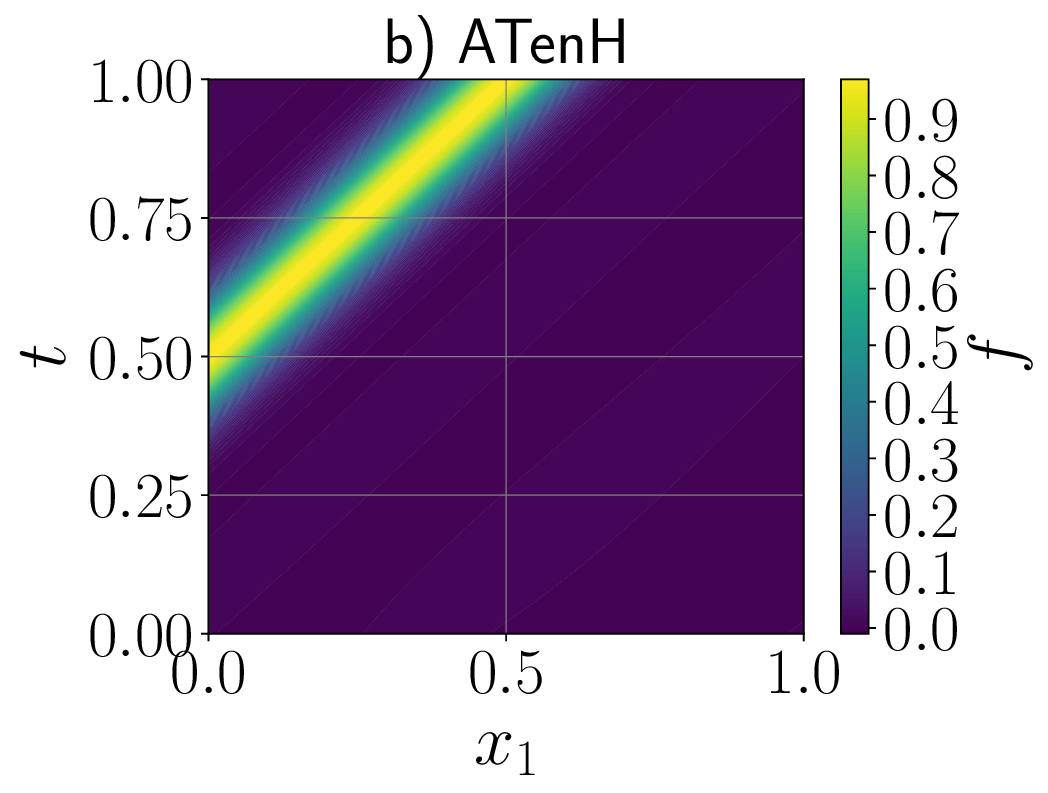} &
        \includegraphics[width=0.33\textwidth]{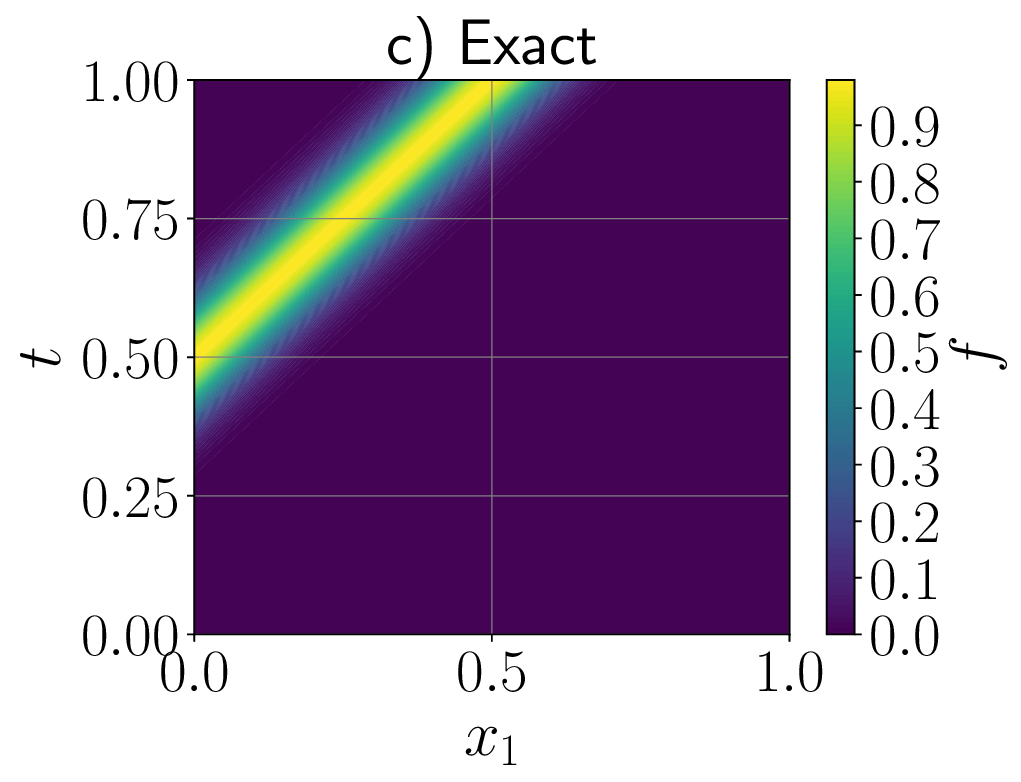} \\

        \includegraphics[width=0.33\textwidth]{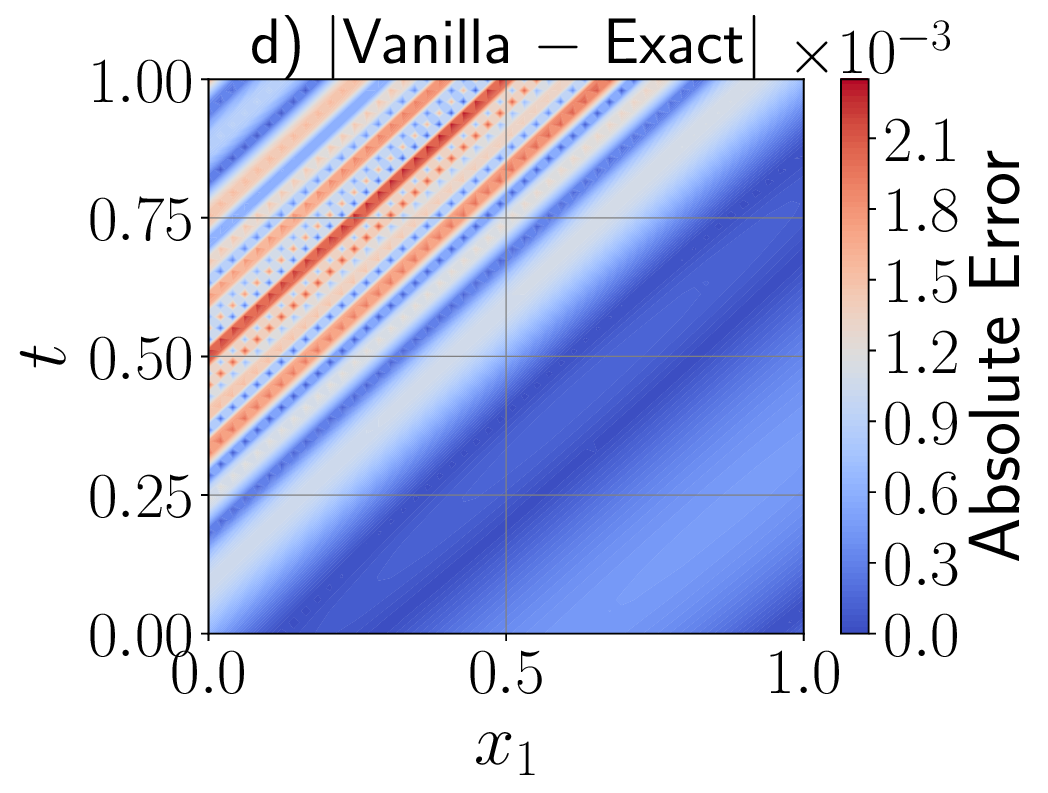} &
        \includegraphics[width=0.33\textwidth]{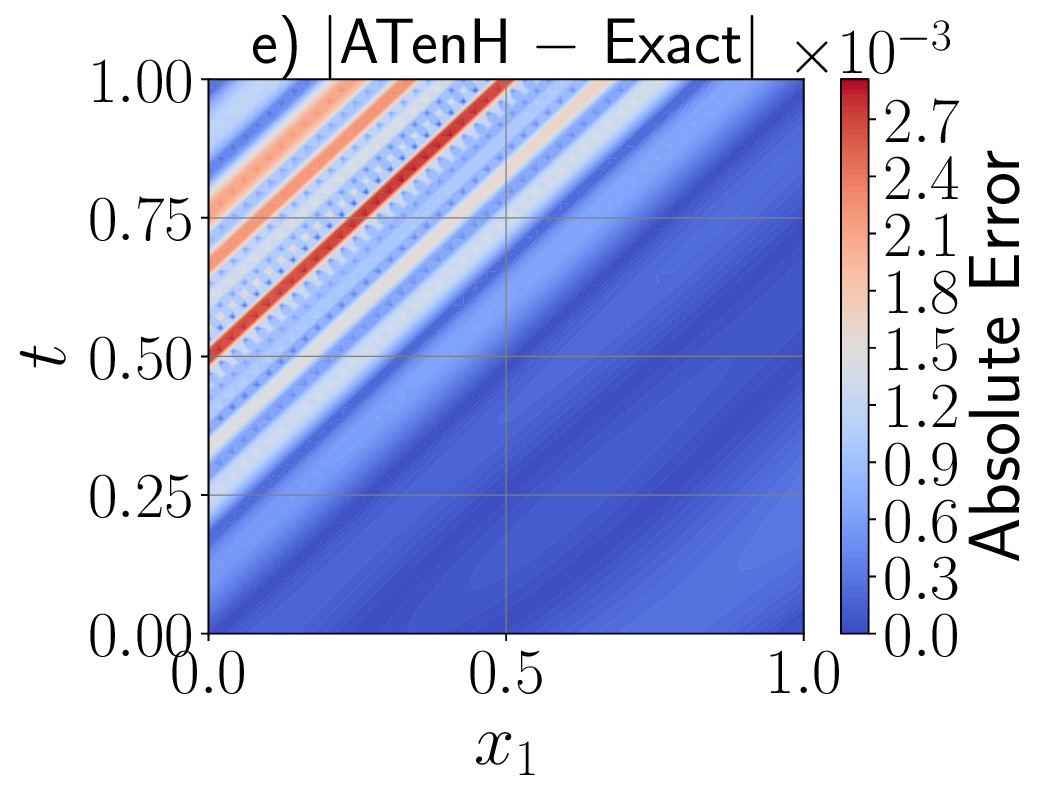} &
        \\ 
    \end{tabular}
    \caption{Representative comparison of the exact solution and the predictions of the Vanilla and ATenH PI-DeepONet architectures for a test input function in 1D linear advection. The models are trained with 10 collocation points per dimension and 500 training input functions. (a) The Vanilla PI-DeepONet's prediction. (b) The ATenH PI-DeepONet's prediction. (c) The exact solution. (d) The absolute error for the Vanilla architecture. (e) The absolute error for the ATenH architecture}
    \label{fig:1DAdvection_Solution}
\end{figure}

We first compare the PI-DeepONet architectures on the 1D linear advection problem. Ten separate Vanilla and ATenH PI-DeepONets are trained over 50,000 epochs to solve this problem in order to report means and assess sensitivity to random parameter initialisation. The solutions produced by a Vanilla and an ATenH PI-DeepONet for a representative test input function are shown in Figure~\ref{fig:1DAdvection_Solution}.

Table~\ref{Table:TrainTime_MRRMSE_Scaling} reports the mean training time and MRRMSE for the 1D linear advection problem. The two architectures exhibit comparable predictive accuracy, with closely matching MRRMSE values and similar standard deviations. The ATenH architecture reduces the mean training time by 12.1\% relative to the Vanilla architecture, corresponding to a $1.14\times$ speed-up. For this case, the theoretical cost derived earlier (Eq.~\ref{eq:trainingtime_scaling_2}) predicts a speed-up scaling of 1.64$\times$.

\subsection{2D Linear Advection}
\label{2D Linear Advection}
\begin{figure}[!htb]
    \centering

    \begin{tabular}{@{}c@{}c@{}c@{}}
        \includegraphics[width=0.3325\textwidth]{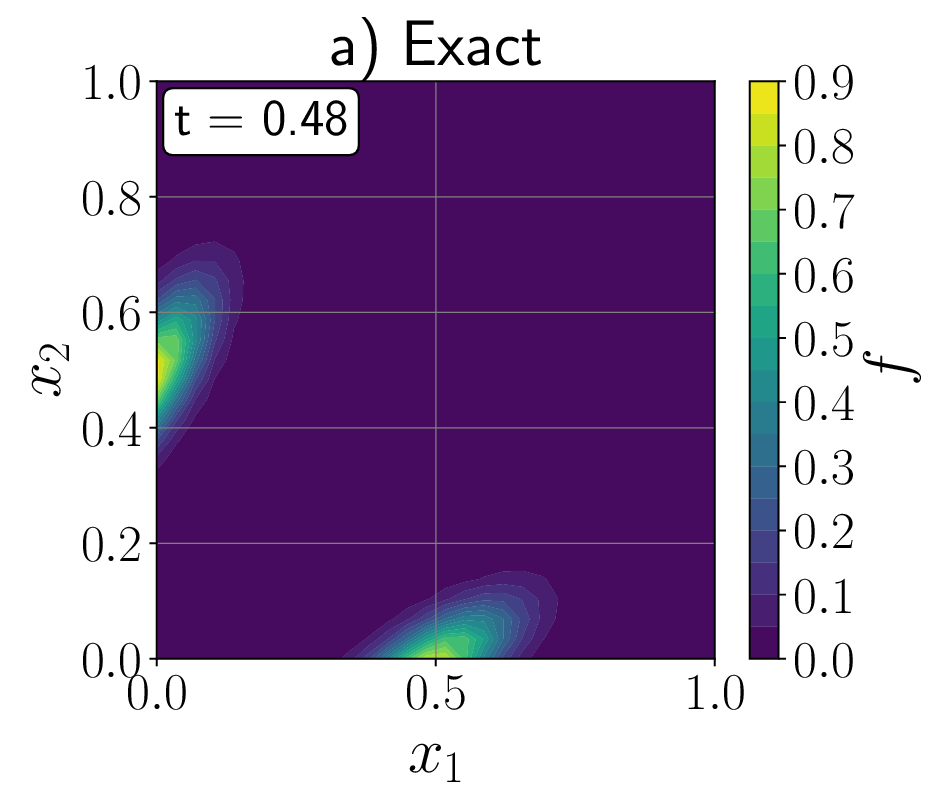} &
        \includegraphics[width=0.3325\textwidth]{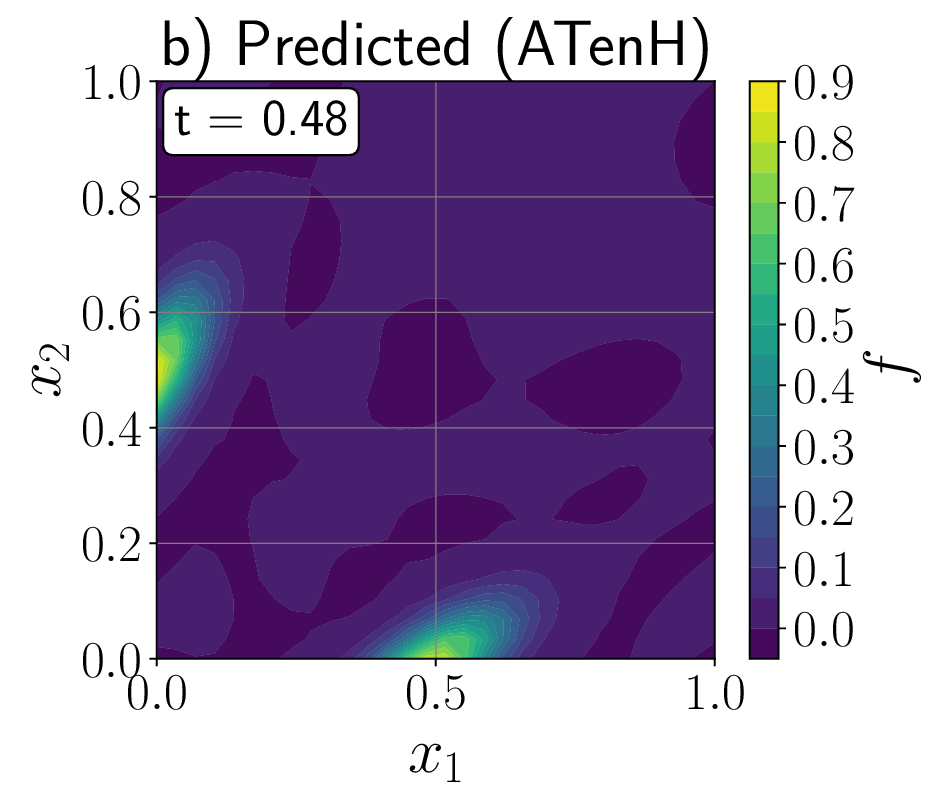} &
        \includegraphics[width=0.325\textwidth]{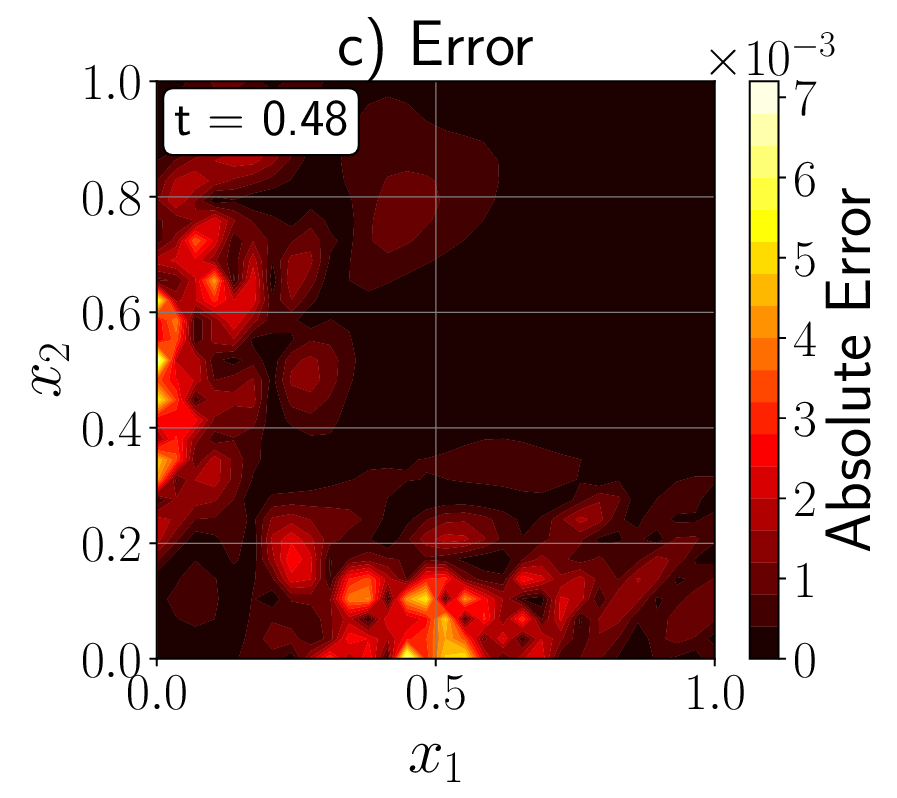} \\

        \includegraphics[width=0.3325\textwidth]{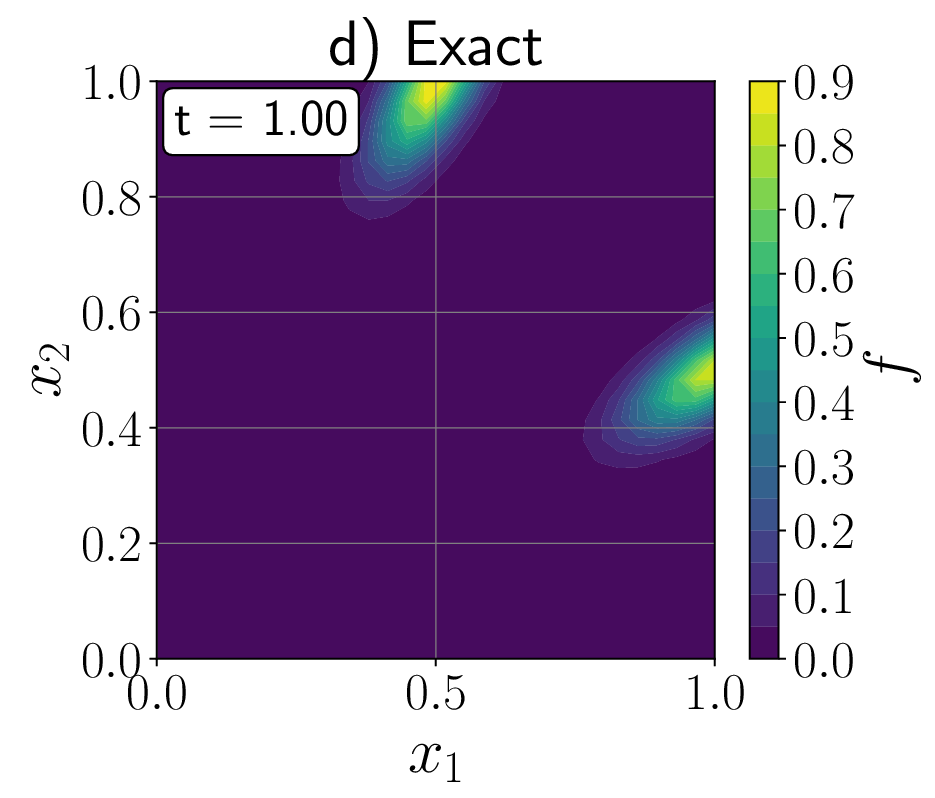} &
        \includegraphics[width=0.3325\textwidth]{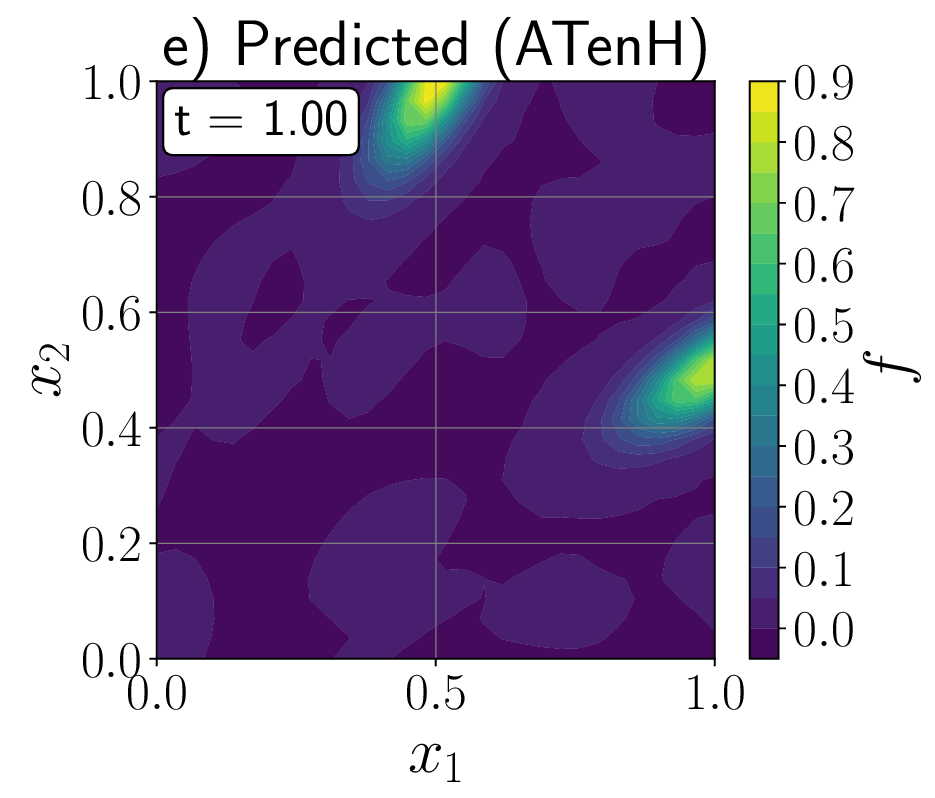} &
        \includegraphics[width=0.325\textwidth]{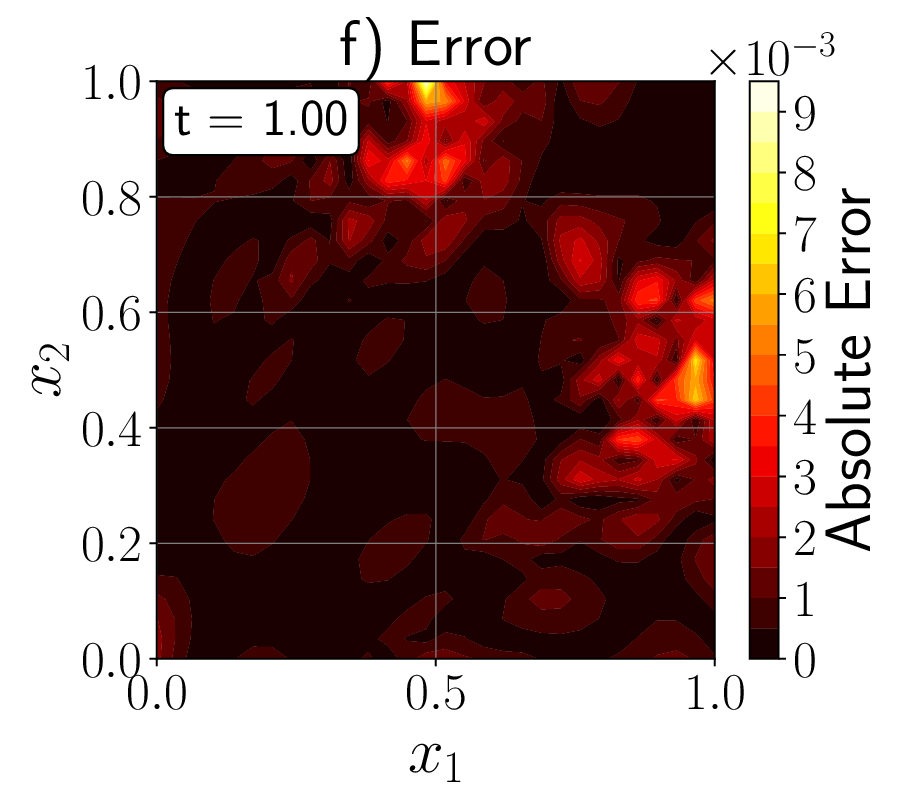} \\ 
    \end{tabular}

    \caption{Comparison of the exact solution and the predictions of an ATenH PI-DeepONet for a representative combination of boundary inputs in 2D linear advection at two time instances. The Vanilla architecture yields qualitatively similar results and is therefore omitted for brevity. (a)--(c) Exact solution, prediction, and absolute error at $t=0.48$. (d)--(f) Exact solution, prediction, and absolute error at $t=1.00$}
    \label{fig:2DAdvection_Solution}
\end{figure}

In this example, the training of Vanilla and ATenH PI-DeepONets are compared for the 2D linear advection problem. The PI-DeepONets are trained for 100,000 full-batch epochs and the average of ten different realisations is reported. An example solution produced by the ATenH implementation of a PI-DeepONet is shown in Fig.~\ref{fig:2DAdvection_Solution}, demonstrating good agreement with the ground truth. The prediction of the Vanilla architecture is omitted due to its close visual similarity to the prediction of the ATenH architecture.

Table~\ref{Table:TrainTime_MRRMSE_Scaling} also reports the mean training time and MRRMSE for linear 2D advection. The two architectures exhibit comparable predictive accuracy, with closely matching MRRMSE values and overlapping standard deviations.

\subsubsection{Computational Performance and Scaling for 2D Linear Advection}
\label{sec:2D_scaling}
Table~\ref{Table:TrainTime_MRRMSE_Scaling} reports the mean training time for linear 2D advection. The ATenH architecture reduces the mean training time by 71.4\%, corresponding to a 3.49$\times$ speed-up. For this case, applying the theoretical cost derived earlier (Eq.~\ref{eq:trainingtime_scaling_2}) with the specified hyperparameters in Section~\ref{Numerical Methodology} predicts a speed-up scaling of 7.46$\times$.

\begin{figure}[!htb]
    \centering

    \begin{minipage}[t]{0.488\textwidth}
        \centering
        \includegraphics[width=\linewidth]{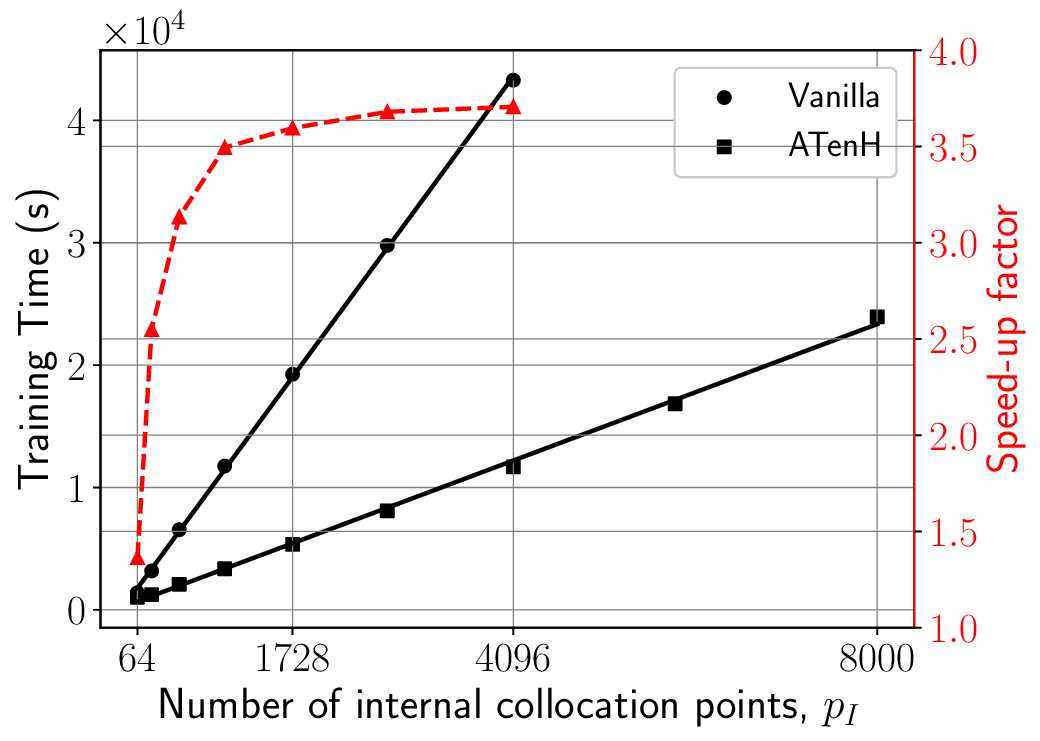}
        \label{fig:2DAdvection_Scaling_collnum}
    \end{minipage}
    \hfill
    \begin{minipage}[t]{0.492\textwidth}
        \centering
        \includegraphics[width=\linewidth]{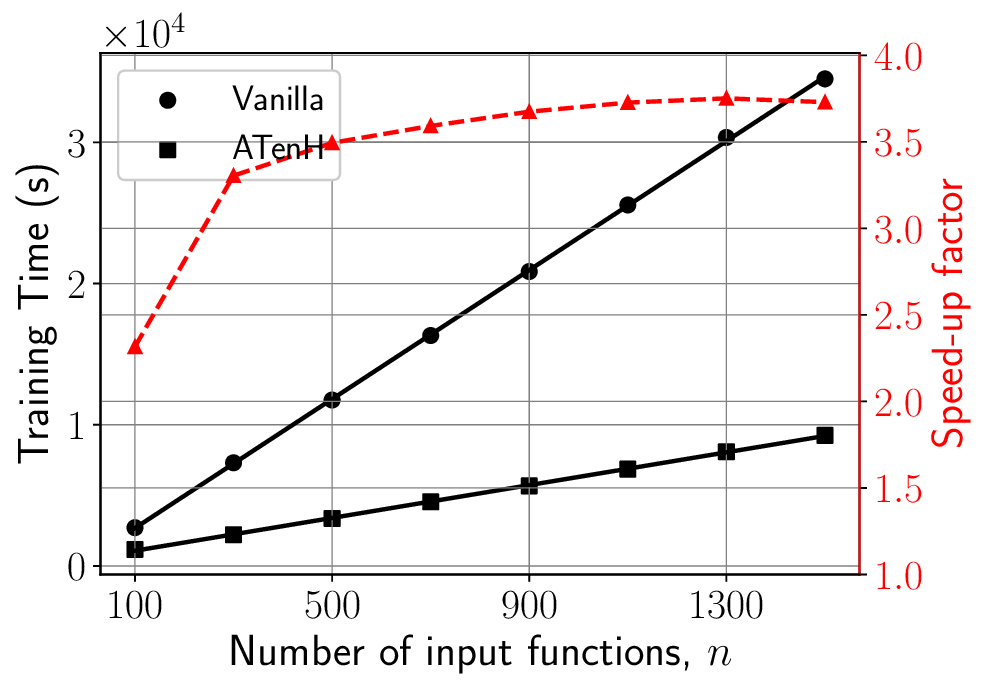}
        \label{fig:2DAdvection_Scaling_funcnum}
    \end{minipage}
    \caption{Mean training time of the Vanilla and ATenH architectures for 2D linear advection with (left) varying numbers of collocation points with a fixed number of input functions, and (right) varying numbers of input functions with a fixed number of collocation points. Solid lines indicate linear regression fits to the training times. Right $y$-axis shows the computed ratio of training times for both architectures (i.e. ATenH's speed-up factor). The two largest collocation-point data points are excluded for the Vanilla architecture due to insufficient memory (left)}
    \label{fig:2DAdvection_Scaling}
\end{figure}

The difference in training computational time between both architectures is more pronounced in 2D relative to 1D. Therefore, we used the 2D problem to analyse the training time scaling for key hyperparameters. Fig.~\ref{fig:2DAdvection_Scaling} shows the mean training time for both architectures for varying numbers of internal collocation points and input functions, respectively, along with the relative speed-up of the ATenH architecture compared to the Vanilla architecture. We use the internal, rather than total, collocation points due to the former being much greater in number than initial/boundary points in practice (specifically, $p_I^{2/3}=p_{b_{x_1}}=p_{b_{x_2}}=p_0$).

For both PI-DeepONet architectures, the training time increases linearly with the number of internal collocation points or input functions. These linear scalings are consistent with the theoretical cost analysis presented in Section~\ref{sec:scaling_whole_step}. In particular, Equations \ref{eq:CaseI_Factoredp_2} and \ref{eq:CaseII_Factoredn_2} show that the computational cost of a training step is proportional to the number of collocation points and the number of input functions, leading to an overall $\mathcal{O}(p)$ and $\mathcal{O}(n)$ dependence. Moreover, the prefactors of these terms are smaller for the ATenH architecture compared to the Vanilla architecture, which explains the consistently lower training times observed for ATenH across all tested collocation point and input function counts.

Using the derived cost expressions in Equation~\ref{eq:trainingtime_scaling_2} with the chosen hyperparameters yields the asymptotic theoretical speed-up factor $\phi \to 7.5$ as $p \to \infty$, while remaining independent of $n$. The value of 7.5 is specific to the chosen values of $m$, $M$, $e$, $L$, $d$, and $s$.

\subsection{3D Linear Advection}
\label{3D Linear Advection}
\begin{figure}[!htb]
    \centering

    \begin{tabular}{@{}c@{}c@{}c@{}}
        \includegraphics[width=0.33\textwidth]{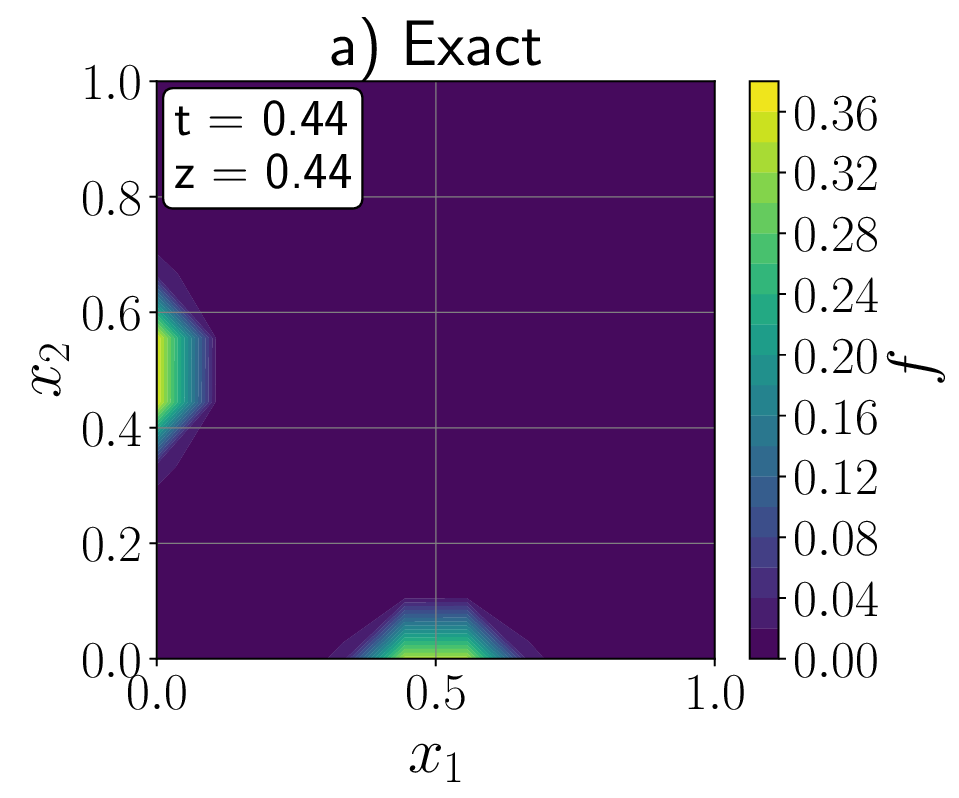} &
        \includegraphics[width=0.33\textwidth]{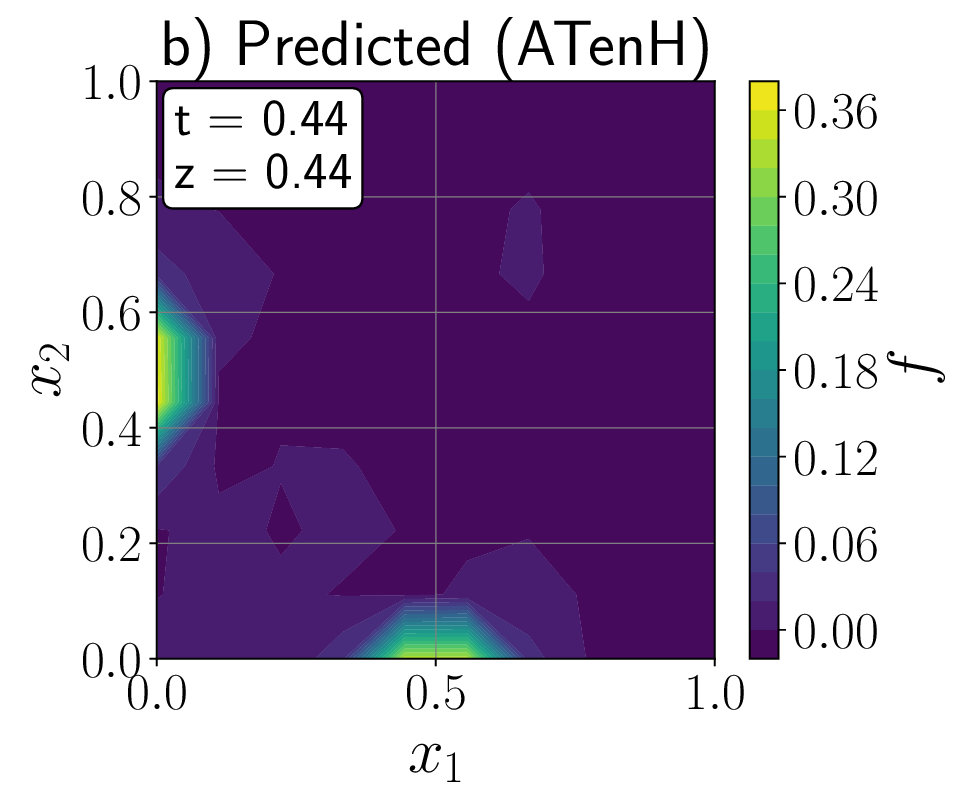} &
        \includegraphics[width=0.33\textwidth]{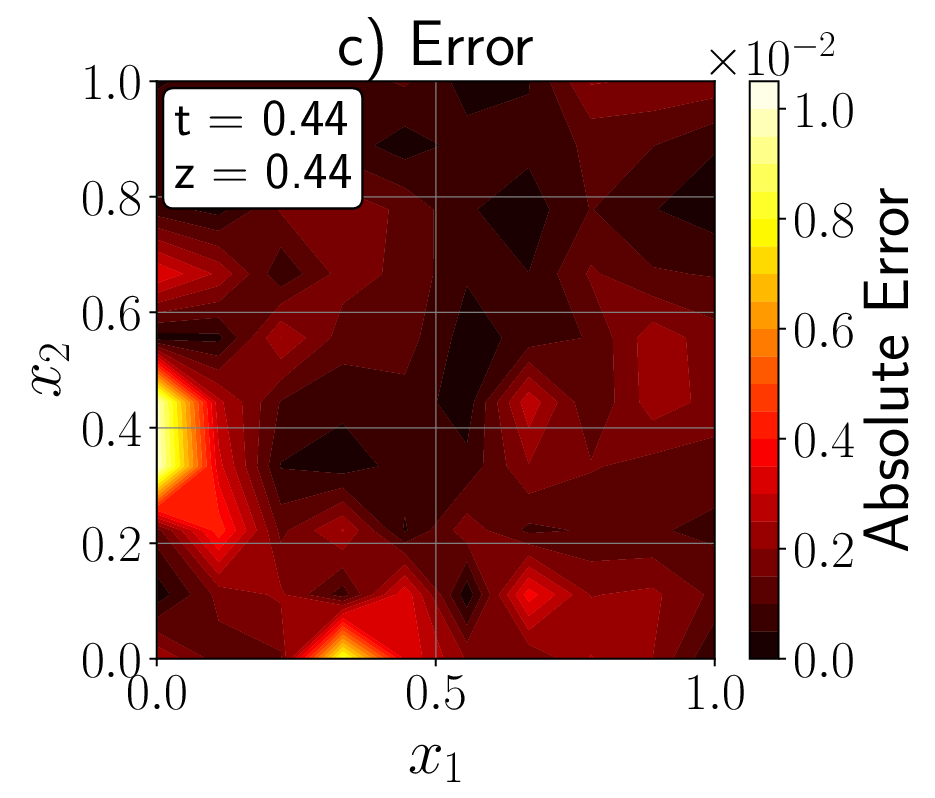} \\

        \includegraphics[width=0.33\textwidth]{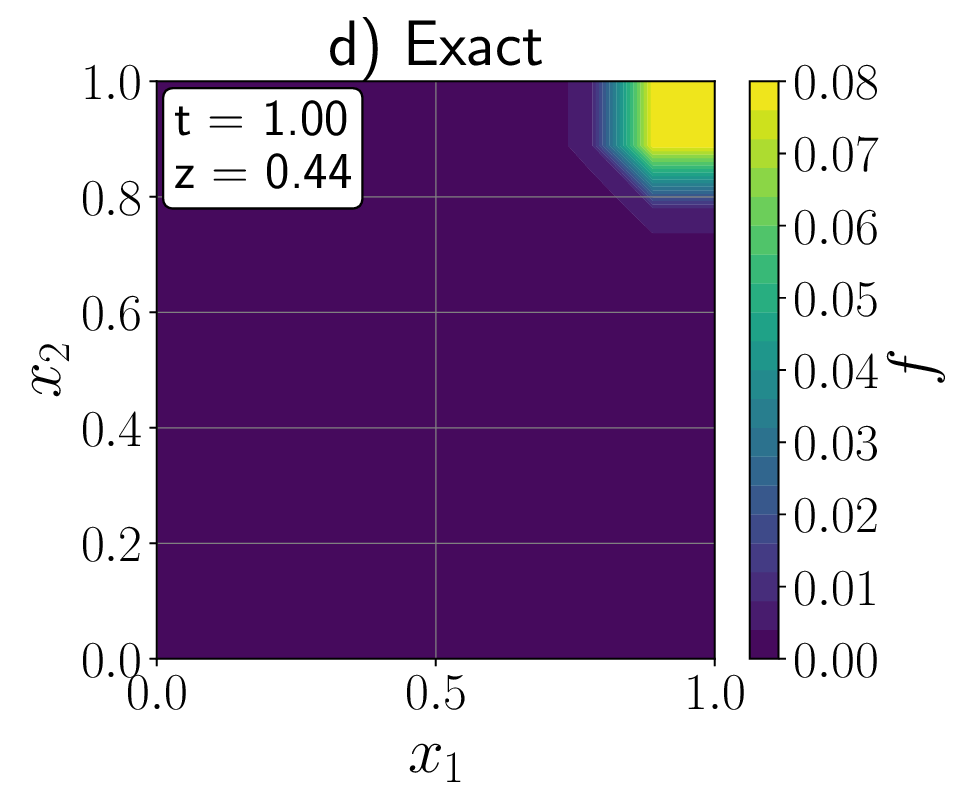} &
        \includegraphics[width=0.33\textwidth]{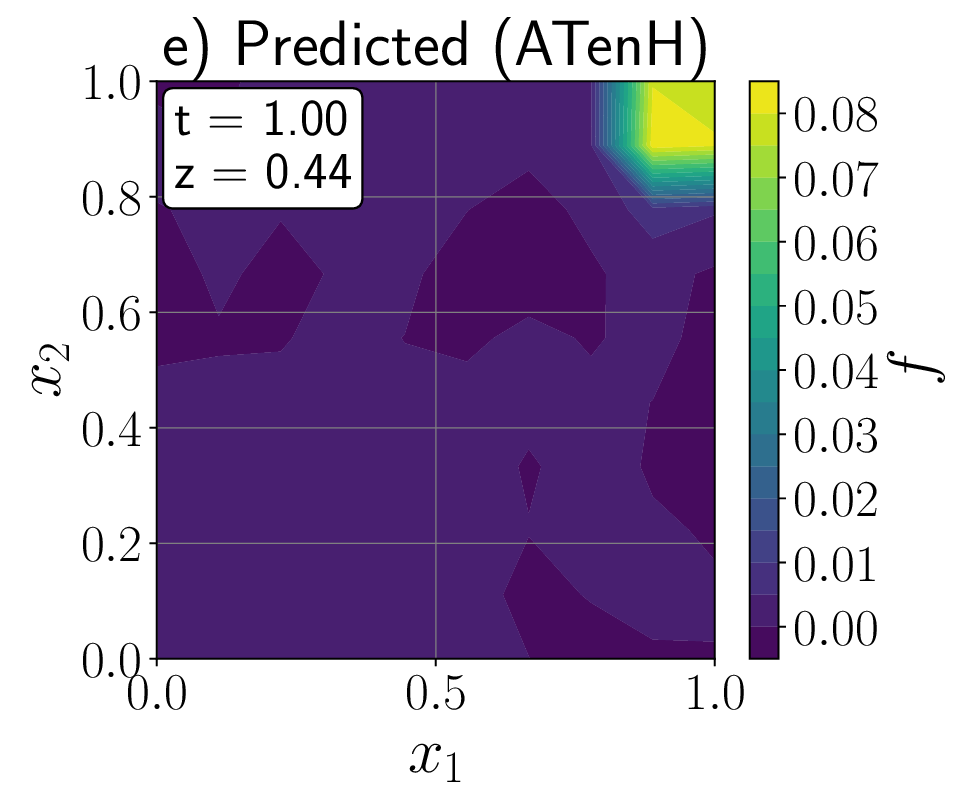} &
        \includegraphics[width=0.33\textwidth]{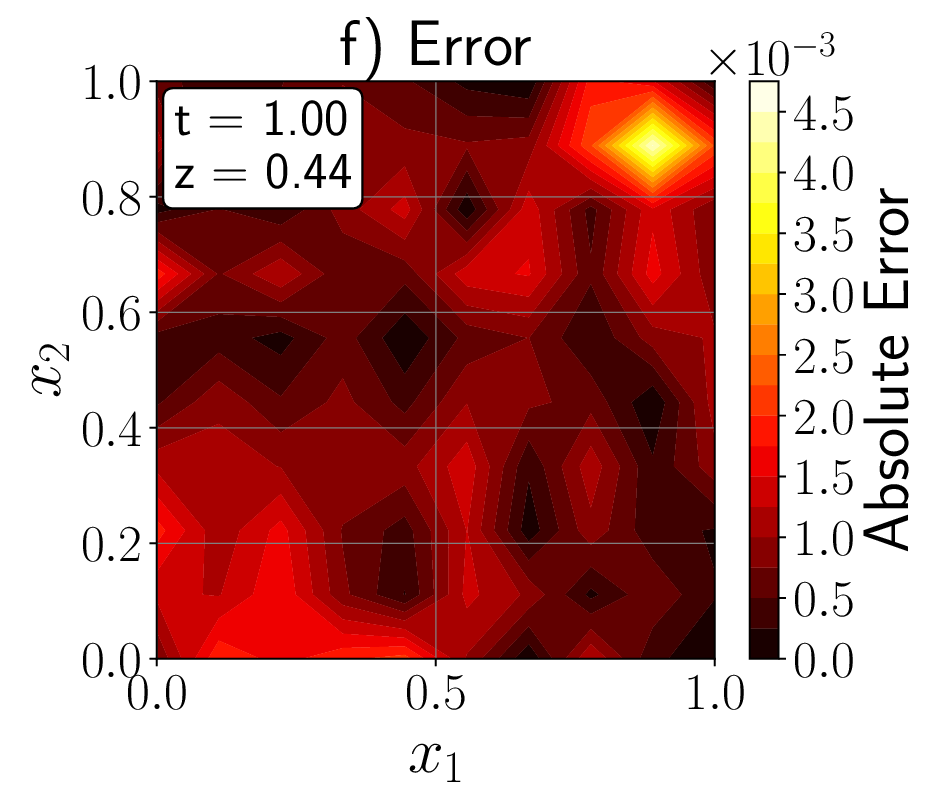} \\ 
    \end{tabular}

    \caption{Comparison of the exact solution and the predictions of an ATenH PI-DeepONet for a representative combination of boundary inputs in 3D linear advection at $x_3=0.44$ and two time instances. The Vanilla architecture fails to train due to a memory error and is therefore omitted. (a)--(c) Exact solution, prediction, and absolute error at $t=0.44$. (d)--(f) Exact solution, prediction, and absolute error at $t=1.00$}
    \label{fig:3DAdvection_Solution}
\end{figure}

In this example, the training of both PI-DeepONet architectures is compared for the 3D linear advection problem. The Vanilla and ATenH PI-DeepONets are trained for 300,000 full-batch epochs and the average of five realisations is reported. Representative exact and PI-DeepONet predicted solutions are shown in Figure~\ref{fig:3DAdvection_Solution}.

Table~\ref{Table:TrainTime_MRRMSE_Scaling} shows that only the ATenH architecture successfully handled this problem, with an average training time of $1.40\times10^5$\,s (approximately 39 hours) and a mean predictive error of $5.438\%$. The Vanilla architecture failed during training due to insufficient memory, highlighting the need to review the implementation of algorithms such as DeepONet, particularly when using widely popular libraries like PyTorch. ATenh succeeds in training here, when the original implementation (Vanilla) fails, thanks to a less memory-intensive input tensor representation. Although an empirical comparison is not possible, Equation~\ref{eq:trainingtime_scaling_2} predicts a theoretical speed-up factor of approximately $\phi\sim580$ for the ATenH architecture relative to the Vanilla architecture.

\begin{table}[ht]
\centering
\caption{Mean training time and MRRMSE scaling with spatial dimensionality in linear advection problems for the Vanilla and ATenH architectures. Standard deviations are not shown for the training times since they are negligible. There is no data for the Vanilla architecture in the 3D linear advection problem due to memory limitations during training. Also shown is the theoretical speed-up $\phi$ for ATenH when compared to the Vanilla PI-DeepONet architecture, for each problem with their corresponding hyperparameters, as specified in Section~\ref{Numerical Methodology}}
\label{Table:TrainTime_MRRMSE_Scaling}
\begin{tabular}{p{1.9cm}ccccc}
\toprule
 & \multicolumn{2}{c}{Vanilla} & \multicolumn{2}{c}{ATenH} & \multicolumn{1}{p{1.8cm}}{Theoretical speed-up} \\
\cmidrule(lr){2-3} \cmidrule(lr){4-5} \cmidrule(lr){6-6}
Spatial dimensionality & Time (s) & MRRMSE (\%) & Time (s) & MRRMSE (\%) & $\phi$ \\
\midrule
1D & $3.14\times10^2$ & $0.471 \pm 0.293$ & $2.76\times10^2$ & $0.308 \pm 0.078$ & 1.64 \\
2D & $1.18\times10^4$ & $2.094 \pm 0.846$ & $3.36\times10^3$ & $1.524 \pm 0.513$ & 7.46 \\
3D & --- & --- & $1.40\times10^5$ & $5.438 \pm 1.108$ & 580 \\
\bottomrule
\end{tabular}
\end{table}

\section{Discussion}
\label{Discussion}
Since the introduction of PI-DeepONets, research has largely focused on improving predictive accuracy and extending the range of problems to which they can be applied. Comparatively less attention has been paid to reducing the computational cost of training especially when scaling PI-DeepONets to computationally demanding, higher-dimensional problems. This work addresses this issue by introducing an architectural modification that preserves the defining advantages of PI-DeepONets – training without labelled solution data, generalisation across varying initial and boundary conditions, and flexibility across PDE classes – while substantially reducing computational cost. 

In this work, we propose a modified PI-DeepONet architecture that changes how input functions are processed within the Branch network. The key observation is that, in the Vanilla implementation of PI-DeepONets, the Branch network is evaluated repeatedly at every collocation point, even though the input function itself does not change. Addressing this redundancy yields our proposed alternative (ATenH), in which the Branch network is evaluated once per input function and its encoding is reused across all collocation points.

This modification comes with the constraint that all output functions must be evaluated at the same number of collocation points. In a data-driven setting this may be restrictive because the amount of labelled training data that is available will likely vary between samples. However, DeepONets are increasingly being adopted within physics-informed frameworks (PI-DeepONets), where this limitation does not apply – collocation points can be freely generated and chosen.

By reducing the dimensionality of the Branch network input tensors, the number of operations required by this subnetwork is substantially decreased. As the Branch network encodes the input functions, which may have a high number of degrees of freedom to adequately represent boundary or initial conditions, this reduction directly lowers the dominant computational cost. Importantly, the modified architecture performs the same underlying mathematical operations as the original one and, as a result, achieves comparable predictive accuracy while significantly reducing training time. For 1D linear advection, the average training time is 314s and 276s for the Vanilla and ATenH architectures, respectively, while for 2D linear advection the corresponding times are 11800s and 3360s. Most notably, the 3D linear advection problem exposed a fundamental practical limitation of the Vanilla PI-DeepONet implementation. It exceeded the available GPU memory and could not be trained on an NVIDIA A100 GPU, whereas the ATenH architecture completed training in approximately 39 hours and achieved an average relative error of 5.44\%. Equation~\ref{eq:trainingtime_scaling_2} also predicts a theoretical speed-up factor of 580$\times$ for this problem. The experimental results demonstrate that the proposed modification does more than reduce training time: by lowering memory requirements, it enables classes of higher-dimensional PI-DeepONet problems that are otherwise computationally infeasible on the same hardware. Consequently, extending PI-DeepONets to higher-dimensional problems is not simply a matter of applying the existing formulation to larger problems; careful architectural design is essential for maintaining computational tractability.

A theoretical cost model was developed to quantify the training-time scaling of both architectures with respect to the number of collocation points and the number of input functions. This model predicts linear scaling with both number of collocation points and input functions for both architectures, but with larger prefactors for the Vanilla architecture, implying steeper scaling. The empirical results for the 1D and 2D linear advection problems confirm this behaviour. The agreement between the theoretical scaling laws and the empirical measurements suggests that the observed performance gains arise primarily from differences in the computational structure of the two architectures rather than implementation-specific effects. The model also predicts an asymptotic constant speed-up factor, which is reinforced by the experimental profiles seen in Figure~\ref{fig:2DAdvection_Scaling}; however, the observed speed-ups are smaller than the theoretical values (1.14$\times$ vs.\ 1.64$\times$ in 1D, and 3.49$\times$ vs.\ 7.46$\times$ in 2D). Despite the quantitative discrepancies, the consistency in qualitative scaling behaviour reinforces the observation that the theoretical model captures the dominant computational trends governing both architectures. Some discrepancy is expected, as the cost model accounts only for the number of summed products in forward and backward passes and therefore neglects implementation-dependent overheads such as memory allocation, activation function evaluation, data movement, automatic differentiation graph construction, and optimiser updates.

\section{Conclusion}
\label{Conclusion}
This work demonstrates that modifying the Branch network input handling in PI-DeepONets can substantially reduce computational cost without sacrificing predictive accuracy. While the theoretical analysis captures the qualitative scaling behaviour, the empirical results quantify substantial practical gains, particularly in higher-dimensional settings.

A key outcome is that the ATenH architecture enables problems that are infeasible under the Vanilla formulation. In particular, while the Vanilla architecture failed on the 3D linear advection problem due to memory limitations, the ATenH architecture successfully completed training with a mean predictive error of 5.44\%. This highlights the fact that implementation-level considerations can determine whether a problem is tractable in practice.

Overall, this work demonstrates that implementation choices can play a critical role in the practical performance of PI-DeepONets. ATenH provides significant improvements over the Vanilla PI-DeepONet formulation; however, it should not be viewed as a final design, but rather as evidence that careful consideration of how these models are implemented can yield substantial gains. This suggests that further improvements may be achievable through continued optimisation of existing architectures, particularly within widely used frameworks such as PyTorch. Furthermore, the agreement between theoretical and empirical scaling indicates that these findings are not tied to a specific hardware or software configuration, but are expected to generalise to other implementations. Future work should explore the application of similar architectural improvements to tackle problems that have so far proved intractable for PI-DeepONets.

\section*{Declarations}

\begin{itemize}
    \item \textbf{Acknowledgements:} The authors acknowledge the use of the IRIDIS High Performance Computing Facility, and associated support services at the University of Southampton, in the completion of this work.
    \item \textbf{Funding:} The first author was supported by UK Research and Innovation (UKRI) through a PhD studentship.
    \item \textbf{Author contributions:} NB: Conceptualisation, Methodology, Software, Visualisation, Analysis, Writing - Original Draft. GA and SM: Conceptualisation, Methodology, Supervision, Writing – Review \& Editing. All authors discussed the results, contributed to the interpretation of the findings, and approved the final manuscript.
    \item \textbf{Competing interests:} The authors declare that they have no conflict of interest.
\end{itemize}





\bibliographystyle{unsrt}  
\bibliography{references}

\end{document}